%% file: Manuscript.tex
\documentclass[3p,oneThankscolumn]{elsarticle}

\makeatletter
\def\ps@pprintTitle{%
 \let\@oddhead\@empty
 \let\@evenhead\@empty
 \def\@oddfoot{\centerline{\thepage}}%
 \let\@evenfoot\@oddfoot}
\makeatother

\usepackage{filecontents}
\RequirePackage{fix-cm}
\usepackage[T1]{fontenc}
\usepackage[utf8]{inputenc}
\usepackage{textcomp}
\usepackage{amsmath}
\usepackage{soul}
\usepackage{mathrsfs}
\usepackage [english]{babel}
\usepackage [autostyle, english = american]{csquotes}
\usepackage{graphicx}
\usepackage{multicol}
\usepackage[table]{xcolor}
\usepackage{collcell}
\usepackage{array, hhline}
\usepackage{pgf}
\usepackage{float}
\usepackage[edges]{forest}
\usetikzlibrary{shadows.blur}
\usepackage[ruled,linesnumbered]{algorithm2e}
\usepackage{subcaption}
\usepackage{amssymb}
\usepackage{xspace}
\usepackage{xtab,afterpage}
\usepackage{multirow}
\usepackage{hyperref}
\newcommand*{\colorme}[1]{%
    \pgfmathparse{#1<.5?1:0}%
    \ifnum\pgfmathresult=0\relax\color{white}\fi
    \pgfmathparse{1-#1}
    \expandafter\cellcolor\expandafter[\expandafter\gray\expandafter]\expandafter{\pgfmathresult}%
    #1%
}

\newcommand{\latinphrase}[1]{\textit{#1}} 
\newcommand{\etal}{\latinphrase{et~al.}\xspace}
\newcommand{\ie}{\latinphrase{i.e.}\xspace}

\newcommand{\eg}{\latinphrase{e.g.}\xspace}
\newcommand{\etc}{\latinphrase{etc.}\xspace}

\begin{document}

\title{Visual Attention Methods in Deep Learning: An In-Depth Survey}
\author[1,2]{Mohammed Hassanin}
\ead{mff00@fayoum.edu.eg}
\author[3,4]{Saeed Anwar\fnref{fn1}}
\ead{saeed.anwar@kfupm.edu.sa}
\author[5]{Ibrahim Radwan}
\ead{ibrahim.radwan@canberra.edu.au}
\author[6,7]{Fahad Shahbaz Khan}
\ead{fahad.khan@mbzuai.ac.ae}
\author[8]{Ajmal Mian}
\ead{ajmal.mian@uwa.edu.au}

\address[1]{University of South Australia,  Australia}
\address[2]{The Faculty of Computers and Information, Fayoum University, Egypt}
\address[3]{King Fahad University of Petroleum and Minerals, Dhahran 31261, KSA}
\address[4]{SDAIA-KFUPM Joint Research Center for Artificial Intelligence (JRCAI), Dhahran, Saudi Arabia}
\address[5]{The University of Canberra, Australia}
\address[6]{Mohamed Bin Zayed University of Artificial Intelligence, Abu Dhabi, UAE}
\address[7]{Computer Vision Laboratory, Linkoping University, Sweden}
\address[8]{The University of Western Australia, Australia}

\fntext[fn1]{Corresponding author.}

\begin{abstract}
\input{abstract.tex}
\end{abstract}

\maketitle
  
\input{intro}
\input{methods}
\input{AttAppType}

\input{challenges}
\input{conclusions}

\bibliographystyle{spmpsci}      
\bibliography{refs}   
\end{document}

%% file: abstract.tex
Inspired by the human cognitive system, attention is a mechanism that imitates the human cognitive awareness about specific information,  
amplifying critical details to focus more on the essential aspects of data. 
Deep learning has employed attention to boost performance for many applications. Interestingly, the same attention design can suit processing different data modalities and can easily be incorporated into large networks. Furthermore, multiple complementary attention mechanisms can be incorporated into one network. Hence, attention techniques have become extremely attractive. 
However, the literature lacks a comprehensive survey on attention techniques to guide researchers in employing attention in their deep models.
Note that, besides being demanding in terms of training data and computational resources, transformers only cover a single category in self-attention out of the many categories available. We fill this gap and provide an in-depth survey of 50 attention techniques, categorizing them by their most prominent features. We initiate our discussion by introducing the fundamental concepts behind the success of the attention mechanism. Next, we furnish some essentials such as the strengths and limitations of each attention category, describe their fundamental building blocks, basic formulations with primary usage, and applications specifically for computer vision. We also discuss the challenges and general open questions related to attention mechanisms. Finally, we recommend possible future research directions for deep attention. All the information about visual attention methods in deep learning is provided at \href{https://github.com/saeed-anwar/VisualAttention}{https://github.com/saeed-anwar/VisualAttention}

%% file: keywords.tex
Attention Mechanisms, Deep Attention, Attention Modules, Attention in Computer Vision and Machine learning.

%% file: intro.tex
\section{Introduction}
Attention has a natural bond with the human cognitive system. According to cognitive science, the human optic nerve receives massive amounts of data, more than it can process. Thus, the human brain weighs the input and pays attention only to the necessary information. With recent developments in machine learning, more specifically, deep learning, and the increasing ability to process large and multiple input data streams, researchers have adopted a similar concept in many domains and formulated various attention mechanisms to improve the performance of deep neural network models in machine translation~\cite{gehring2017convolutional, m_transformers}, visual recognition~\cite{m_nonlocal}, generative models~\cite{zhang2019self}, multi-agent reinforcement learning~\cite{iqbal2019actor}, \etc Over the past decade, deep learning has advanced in leaps and bounds, leading to many deep neural network architectures capable of 
learning complex relationships in data. Generally, neural networks provide implicit attention to extract meaningful information from the data.

\begin{figure}[t]
\begin{center}
\begin{tabular}{ccc}
\includegraphics[width=0.34\textwidth]{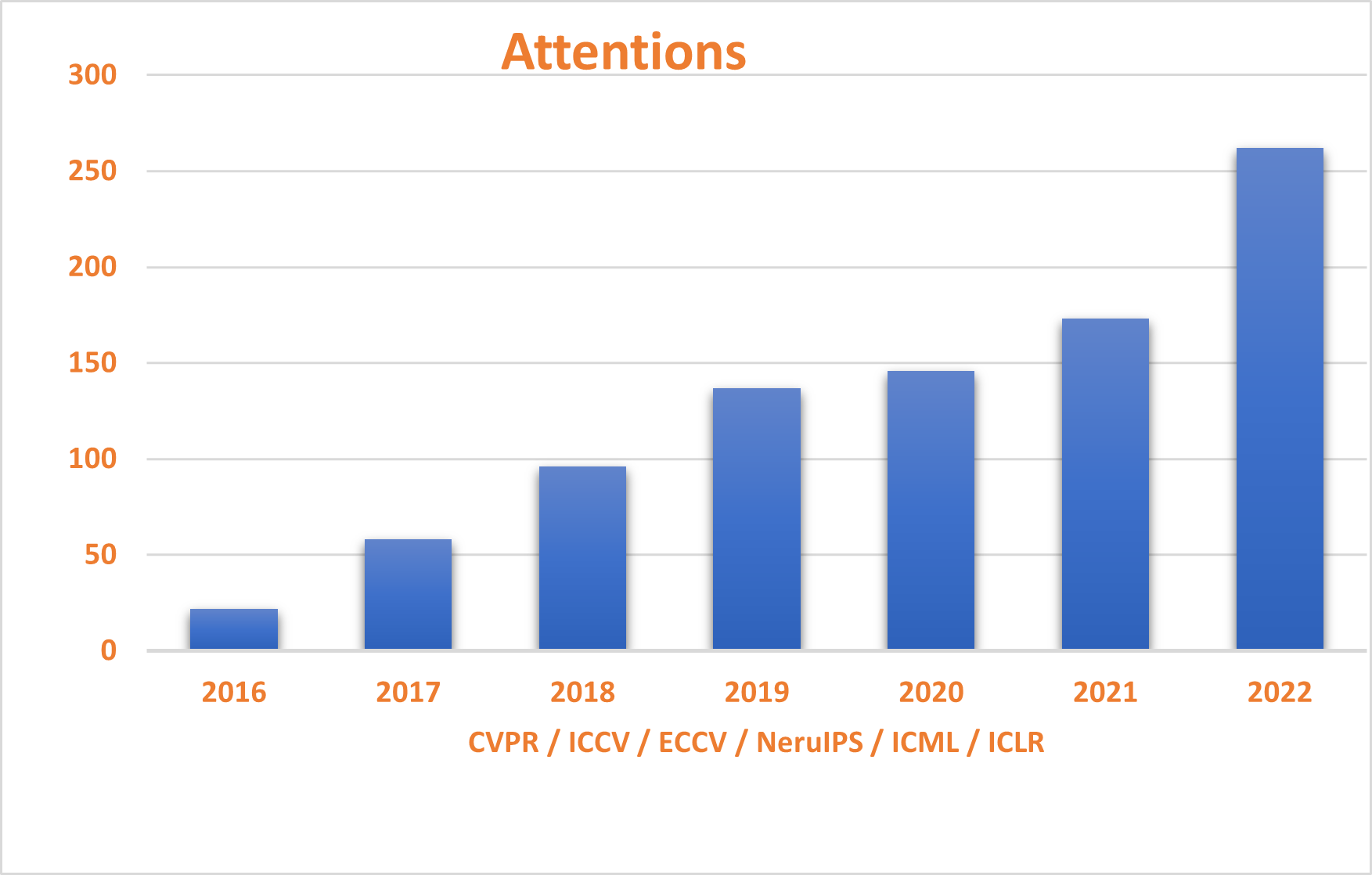}&
\includegraphics[width=0.33\textwidth]{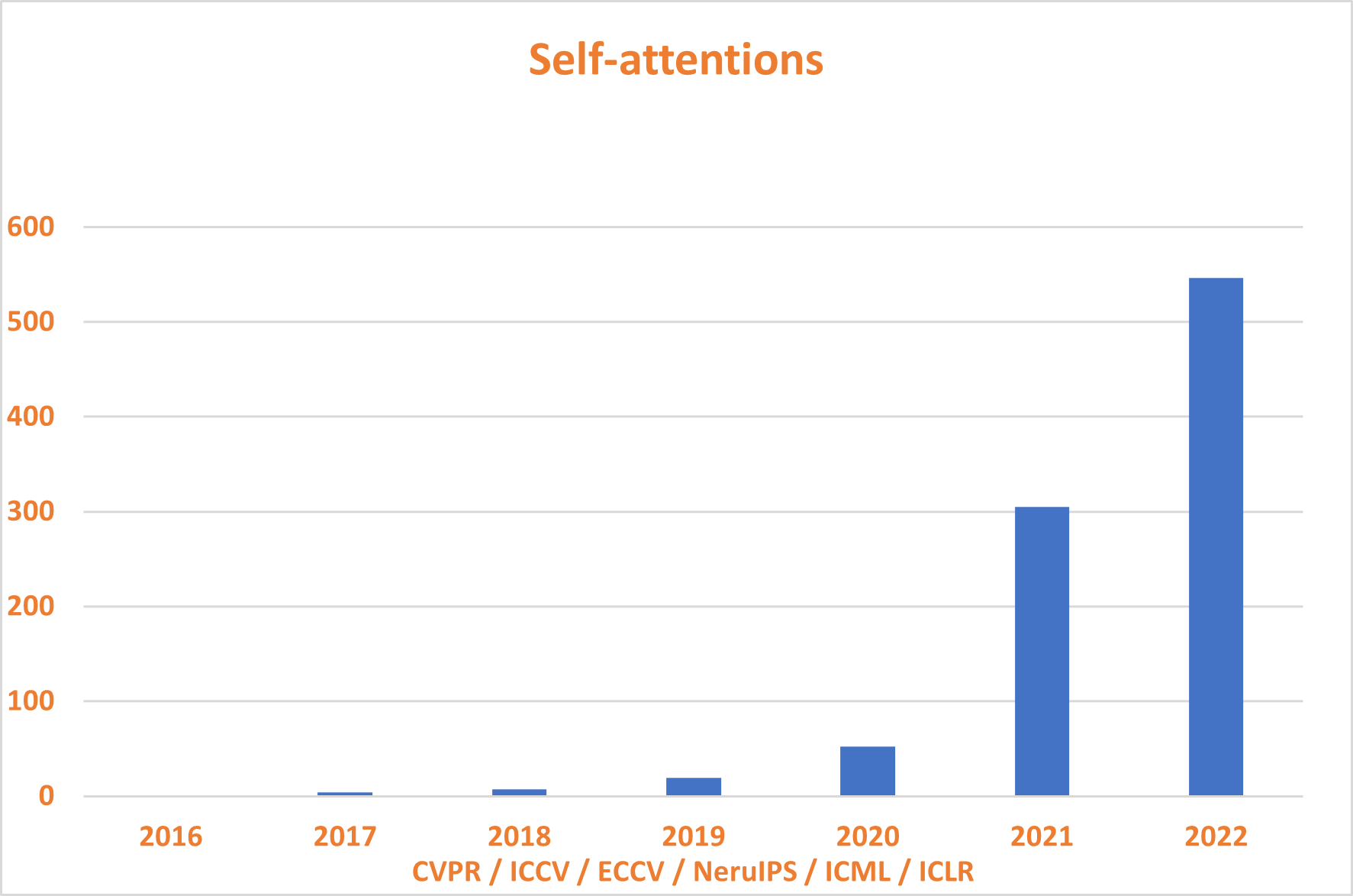}&
\includegraphics[width=0.32\textwidth]{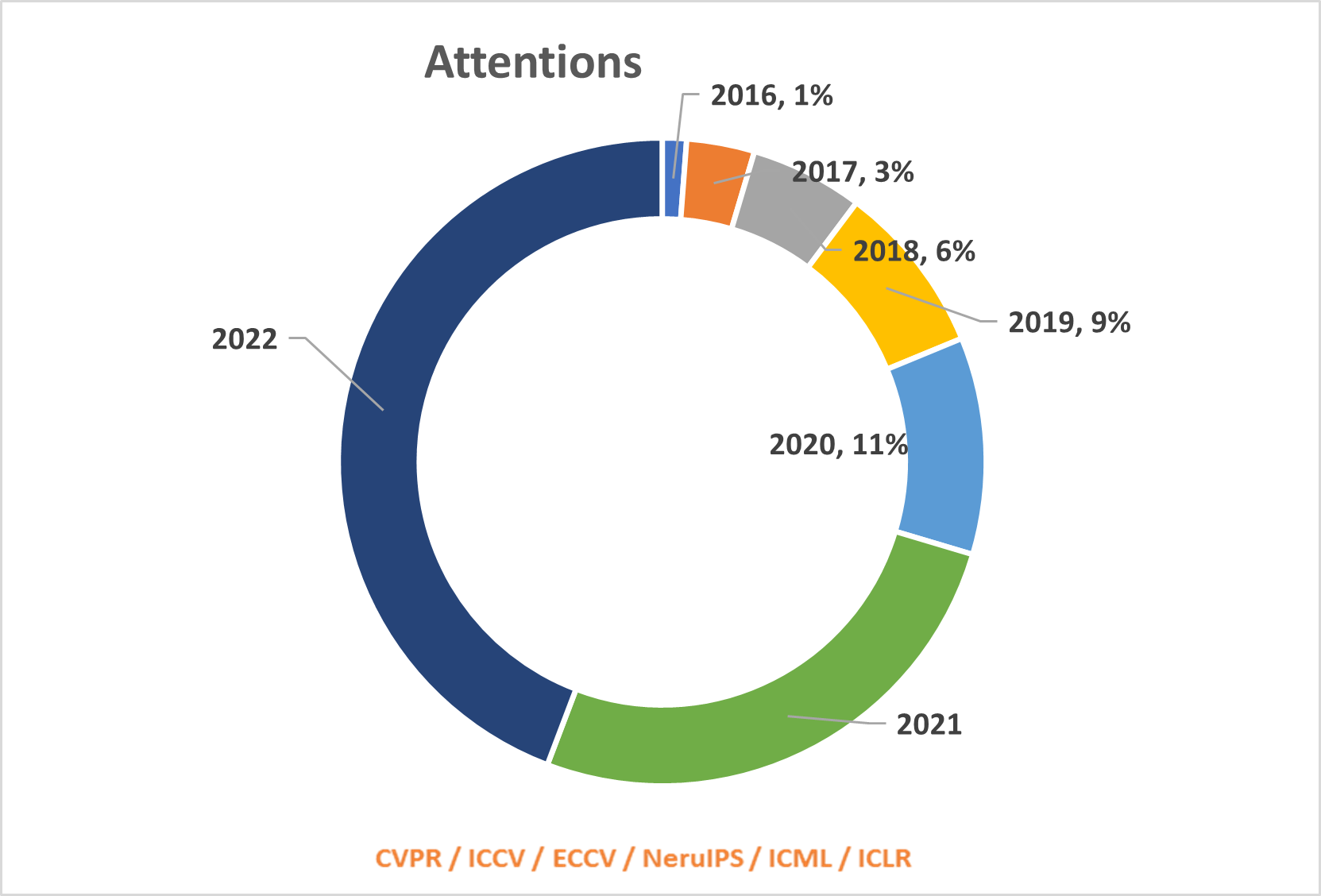}\\
Non-self attentions&
Self-attention methods&
All types of attentions\\
\end{tabular}
\end{center}
\caption{Visual charts show the increase in the number of attention-related papers in the top conferences and journals.}
\label{fig:attentionincrease}
\end{figure}

An explicit attention mechanism in deep learning was first introduced to tackle the \emph{forgetting} issue in encoder-decoder architectures designed for the machine translation problem~\cite{bahdanau2014neural}. Since the network's encoder part focuses on generating a representative input vector, the decoder generates the output from the representation vector. A bi-directional Recurrent Neural Network (RNN)~\cite{bahdanau2014neural} is employed for solving the \emph{forgetting} issue by generating a context vector from the input sequence and then decoding the output based on the context vector as well as the previous hidden states. The context vector is computed by a weighted sum of the intermediate representations
which makes this method an example of explicit attention. 
Moreover, Long-Short-Term-Memory (LSTM)~\cite{sutskever2014sequence} is employed to generate both the context vector and the output. Both methods compute the context vector considering all the hidden states of the encoder. However,~\cite{luong2015effective} introduced another idea by getting the attention mechanism to focus on only a subset of the hidden states to generate every item in the context vector. This was computationally less expensive compared to the previous attention methods and showed a trade-off between \emph{global} and \emph{local} attention mechanisms. 

Another attention-based breakthrough was made by Vaswani~\etal~\cite{m_transformers}, where an entire architecture was created based on the self-attention mechanism. The items in the input sequence are encoded in parallel into multiple representations called key, query, and value. This architecture, coined the Transformer, helps capture the importance of each item relative to others in the input sequence more effectively. Recently, many researchers have extended the basic Transformer architecture for specific applications.

To pay attention to the significant parts in an image and suppress unnecessary information, advancements of attention-based learning have found their way into multiple computer vision tasks, either employing a different attention map for every image pixel, comparing it with the representations of other pixels~\cite{m_nonlocal, dosovitskiy2020image, zhang2019self} or generating an attention map to extract the global representation for the whole image~\cite{kosiorek2017hierarchical, jetley2018learn}. However, the design of the attention mechanism is highly dependent on the problem at hand. To enforce the selection of hidden states that correspond to the critical information in the input, attention techniques have been used as plug-in units in vision-based tasks, alleviating the risk of vanishing gradients. To sum up, attention scores are calculated, and hidden states are selected either deterministically or stochastically. 

\input{TaxonomyFigure2}

Attention has been the center of significant research efforts over the past few years, and image attention has been flourishing in many different machine learning and vision applications, for example, classification~\cite{show_attend}, detection~\cite{zhao2019object}, image captioning~\cite{hossain2019comprehensive}, 3D analysis~\cite{Qiu2021Att3Ddetection, hassanin2022crossformer}, \etc Despite the impressive performance of attention techniques employed in deep learning, no literature survey comprehensively reviews all, especially deep learning based, attention mechanisms in vision to categorize them based on their basic underlying architectures and highlight their strengths and weaknesses.
Recently, researchers surveyed application-specific attention techniques with an emphasis on NLP-based~\cite{hu2019introductory}, transformer-based~\cite{han2020survey, khan2021transformers}, and graph-based approaches~\cite{lee2019attention}. 
However, no comprehensive study collates with the huge and diverse scope of {\em all} deep learning-based attention techniques developed for visual inputs.

In this article, we review attention techniques specific to vision. Our survey covers the numerous basic building blocks (operations and functions) and complete architectures designed to learn suitable representations while making the models attentive to the relevant and important information in the input images or videos. Our survey broadly classifies attention mechanisms proposed in the computer vision literature, including soft attention, hard attention, multi-modal, arithmetic, class attention, and logical attention. We note that some methods belong to more than one category; however, we assign each technique to the category with a dominant association with other methods of that category. Following such a categorization helps track the common attention mechanism characteristics and offers insights that can potentially help design novel attention techniques. Figure~\ref{fig:taxonomy} shows the classification of the attention mechanisms. 
In contrast to the other works that surveyed a specific branch of attention methods, our survey provides a high overview of the different attention techniques. The primary motivation is to provide various attention methods in one paper so that the research community can inspire thoughts and solutions to issues of each one of them. The contributions of this survey are.
\begin{itemize}
    \item The first survey to provide a high overview of attention methods rather than a specific branch.
    \item Providing a categorization to the various attention methods.
    \item Surveying 70 attention methods.
    \item Discussing the challenges and future directions of attention methods.
\end{itemize}

We emphasize that a survey is warranted for attention in vision due to the large number of papers published as outlined in Figure~\ref{fig:attentionincrease}. It is evident from Figure~\ref{fig:attentionincrease} that the number of articles published in the last year has significantly increased compared to previous years, and we expect to see a similar trend in the coming years. Furthermore, our survey lists articles of significant importance to assist the computer vision and machine learning community in adopting the most suitable attention mechanisms in their models and avoiding duplicating attention methodologies. It also identifies research gaps, provides the current research context, and presents plausible research directions and future focus areas.

Since transformers have been employed across many vision applications, a few surveys \cite{khan2021transformers,han2020survey} summarize the recent trends of transformers in computer vision. Although transformers offer high accuracy, this comes at the cost of very high computational complexity, which hinders their feasibility for mobile and embedded system applications. Furthermore, transformer-based models require substantially more training data than CNNs and lack efficient hardware designs and generalizability. According to our survey, transformers only cover a single category in self-attention out of the 50 different attention categories surveyed. Another significant difference is that our survey focuses on attention types rather than applications covered in transformer-based surveys \cite{khan2021transformers,han2020survey}. Additionally, there is a recently published review on attention mechanisms in computer vision~\cite{guo2022attention}, albeit with a limited scope, where it primarily concentrates on limited and basic attention mechanisms such as spatial, temporal, channel, and branch attention methods. In contrast, our survey offers a more comprehensive and technically detailed exploration of various attention mechanisms in vision, where multiple articles that share a similar attention mechanism are summarized and discussed.

%% file: TaxonomyFigure2.tex
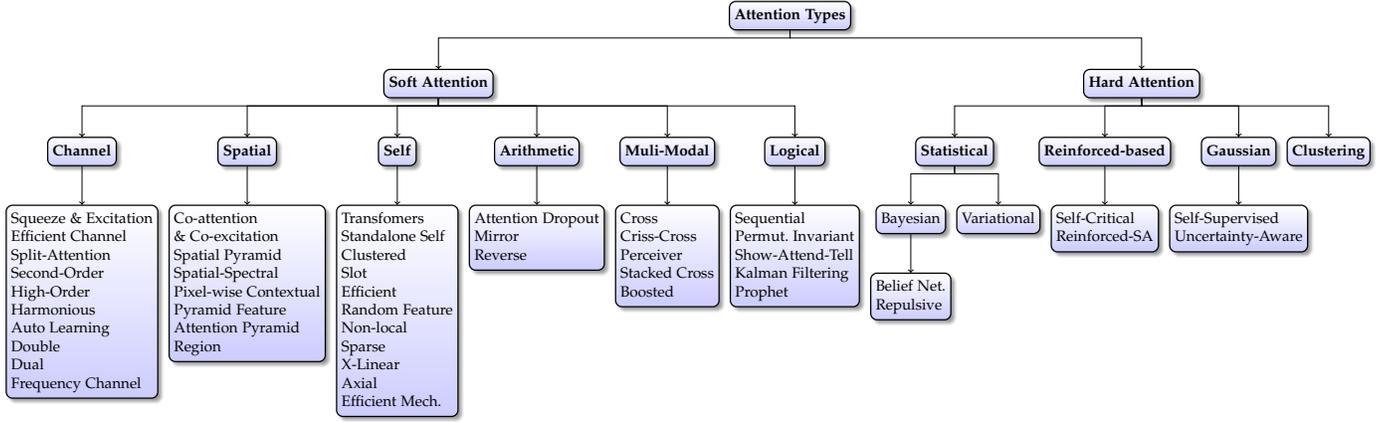
\begin{figure*}[t]
	\centering
	\resizebox{\textwidth}{!}{
	\begin{forest}
		forked edges,
		for tree={thick,draw,align=left,edge={-latex},fill=white,blur shadow, rounded corners, top color=white, bottom color=blue!20,     edge+=->,
        l sep'+=13pt,     
        }
        [\textbf{Attention Types}
        [\textbf{Soft Attention}
		  [\textbf{Channel}
		    [Squeeze $\&$ Excitation\\
		    Efficient Channel\\
		    Split-Attention\\
		    Second-Order \\
		    High-Order\\
		    Harmonious\\
		    Auto Learning\\
		    Double\\
		    Dual\\
		    Frequency Channel
		    ]
		  ]
  		  [\textbf{Spatial}
  		    [Co-attention \\$\&$ Co-excitation\\
  		    Spatial Pyramid\\
  		    Spatial-Spectral\\
  		    Pixel-wise Contextual\\
  		    Pyramid Feature\\
  		    Attention Pyramid\\
  		    Region ]
  		  ]
		  [\textbf{Self}
		    [Transfomers\\
		    Standalone Self\\
		    Clustered\\
		    Slot \\ 
		    Efficient \\
		    Random Feature \\
		    Non-local\\
		    Sparse \\
		    X-Linear \\
		    Axial\\
		    Efficient Mech.]
		   ]
		  [\textbf{Arithmetic}
		    [Attention Dropout\\
		    Mirror \\
		    Reverse ]
		  ]
		  [\textbf{Multi-Modal}
		    [Cross\\
		    Criss-Cross \\
		    Perceiver\\
		    Stacked Cross \\
		    Boosted ]
		  ]
		  [\textbf{Logical}
		    [Sequential \\
		    Permut. Invariant \\
		    Show-Attend-Tell\\
		    Kalman Filtering \\
		    Prophet ]
		  ]
		]
		[\textbf{Hard Attention}
		  [\textbf{Statistical}
		    [Bayesian
		        [Belief Net.\\
		        Repulsive ]
		    ]
		    [Variational]
		  ]
		  [\textbf{Reinforced-based} 
		    [Self-Critical\\
		    Reinforced-SA]
		  ]
		[\textbf{Gaussian}
		   [Self-Supervised\\
		   Uncertainty-Aware]
		]
		[\textbf{Clustering}]
		]
	]
	\end{forest}}
\vspace{1mm}
\caption{A taxonomy of attention types. The attentions are categorized based on the methodology adopted to perform attention. Some attention techniques can be accommodated in multiple categories; in this case, the attention is grouped based on the most dominant characteristic and primary application.}
\label{fig:taxonomy}
\end{figure*}

%% file: methods.tex
\section{Attention in Vision}

The primary purpose of attention in vision is to imitate the human visual cognitive system and focus on the essential features~\cite{hermann2015teaching} in the input image. We categorize attention methods based on the main function used to generate attention scores, such as softmax or sigmoid. Table~\ref{tab:overall} provides the summary, application, strengths, and limitations for the category presented in this survey.

\subsection{Soft (Deterministic) Attention}
This section reviews soft-attention methods such as channel attention, spatial attention, and self-attention. In channel attention, the scores are calculated channel-wise because each one of the feature maps (channels) attends to specific parts of the input. In spatial attention, the main idea is to attend to the critical regions in the image. Attending over regions of interest facilitates object detection, semantic segmentation, and person re-identification. In contrast to channel attention, spatial attention attends to the essential parts of the spatial map (bounded by width and height). It can be used independently or as a complementary mechanism to channel attention. On the other hand, self-attention is proposed to encode higher-order interactions and contextual information by extracting the relationships between input sequence tokens. 
It differs from channel attention in how it generates the attention scores, as it mainly calculates the similarity between two maps (K, Q) of the same input, whereas channel attention generates the scores from a single map. However, self-attention and channel attention both operate on channels. 
Soft attention methods calculate the attention scores as the weighted sum of all the input entities~\cite{luong2015effective} and mainly use soft functions such as softmax and sigmoid. Since these methods are differentiable, they can be trained through back-propagation techniques. However, they suffer from other issues, such as high computational complexity and assigning weights to non-attended objects.

\begin{figure}[t]
  \centering
  \begin{tabular}[b]{c}
    \includegraphics[width=.2\paperwidth, height=.05\paperheight]{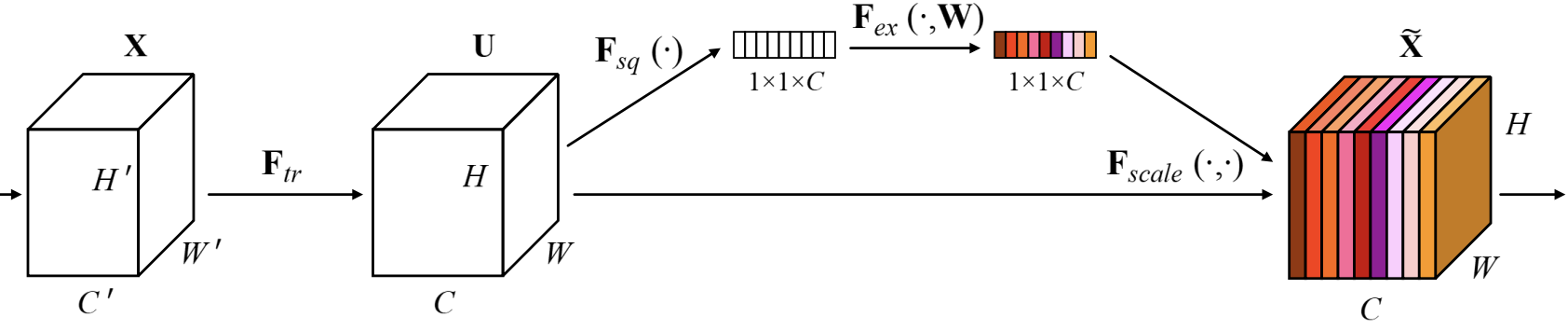} \\
    \small (a) SENet  \cite{hu2018squeeze}
  \tabularnewline
    \includegraphics[width=.22\paperwidth, height=.05\paperheight]{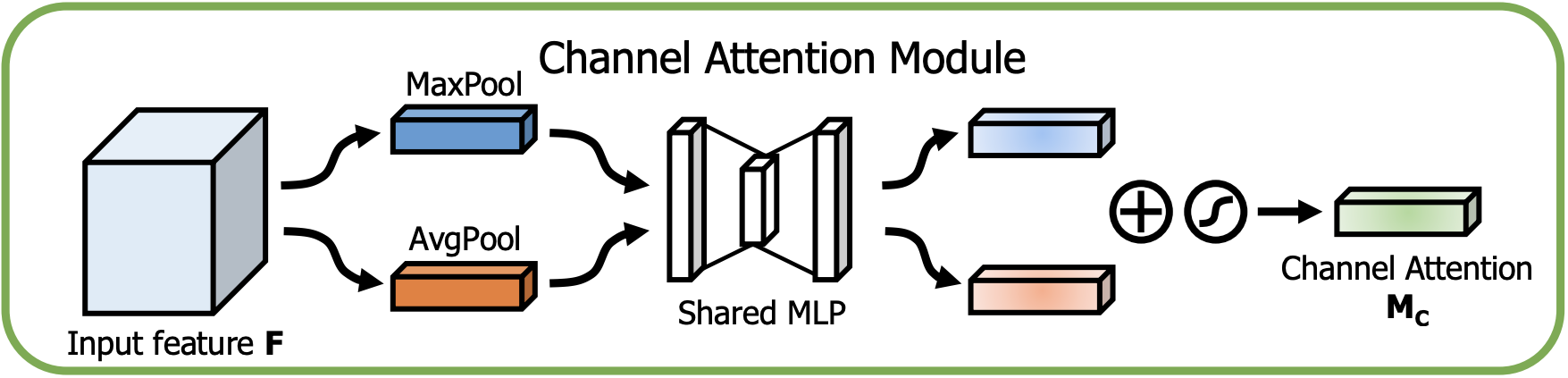} \\
    \small (b) CBAM \cite{woo2018cbam}
  \tabularnewline
    \includegraphics[width=.22\paperwidth, height=.08\paperheight]{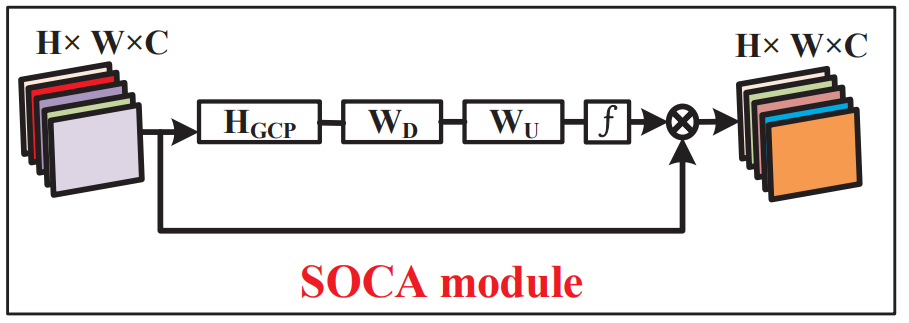} \\
    \small (c) SOCA \cite{Dai_2019_CVPR}
  \tabularnewline
    \includegraphics[width=.22\paperwidth, height=.08\paperheight]{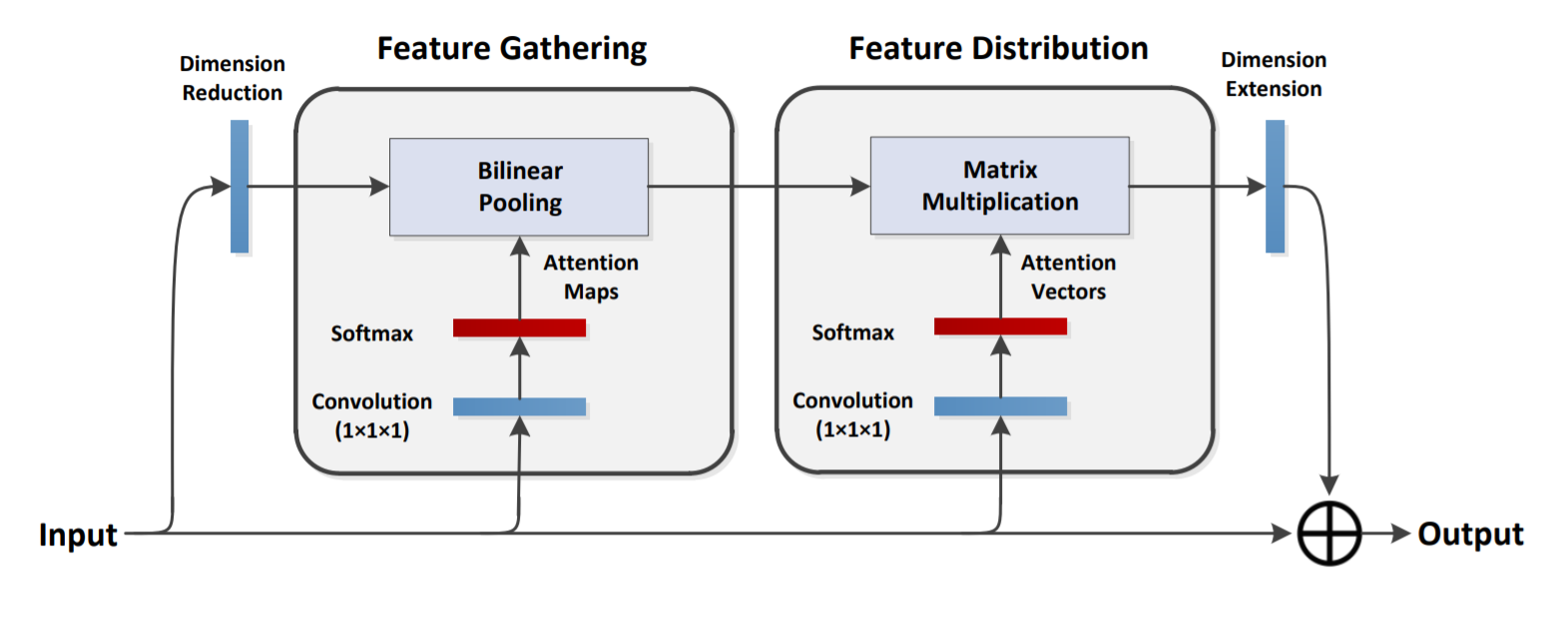} \\
    \small (d) A$^2-$Net \cite{chen20182}
  \end{tabular}
   \begin{tabular}[b]{c}
    \includegraphics[width=.27\paperwidth, height=.09\paperheight]{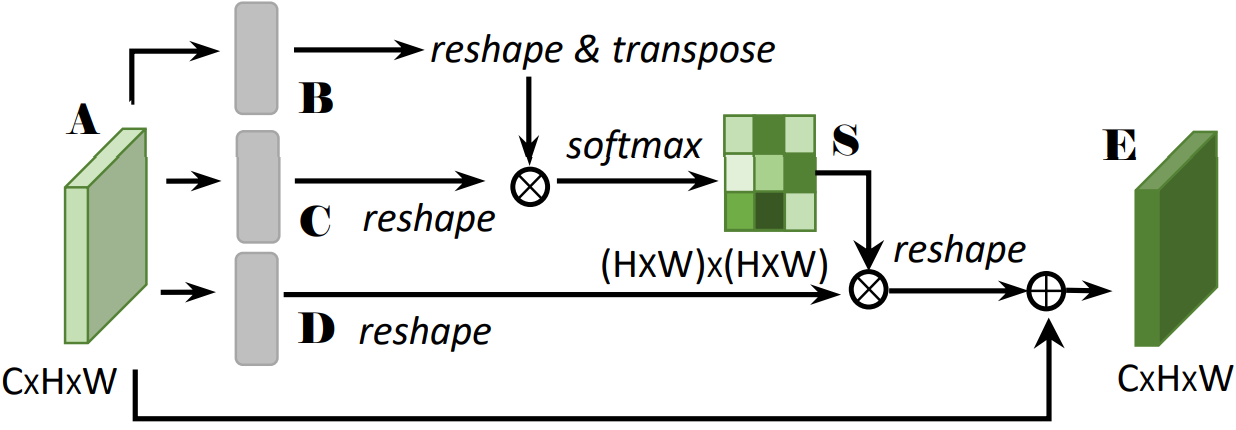} \\
    \small (e) DAN Positional\cite{fu2019dual}
  \tabularnewline
    \includegraphics[width=.27\paperwidth, height=.09\paperheight]{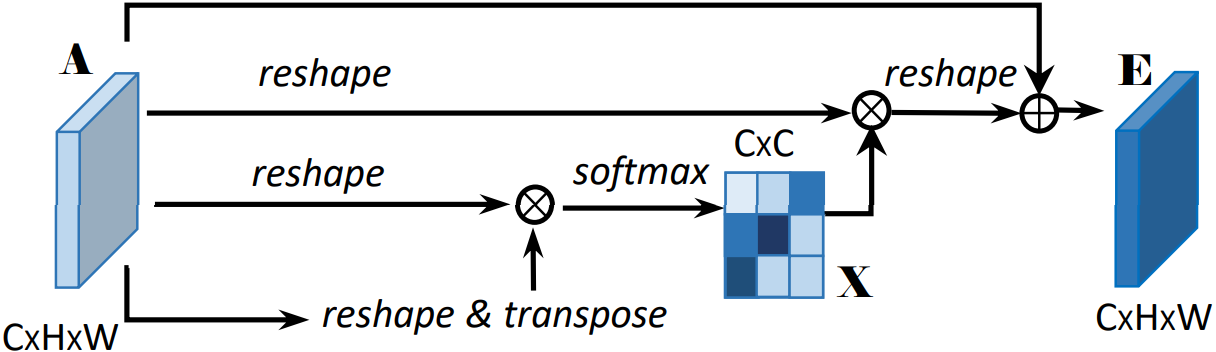} \\
    \small (f) DAN Channel \cite{fu2019dual}
  \tabularnewline
    \includegraphics[width=.25\paperwidth, height=.09\paperheight]{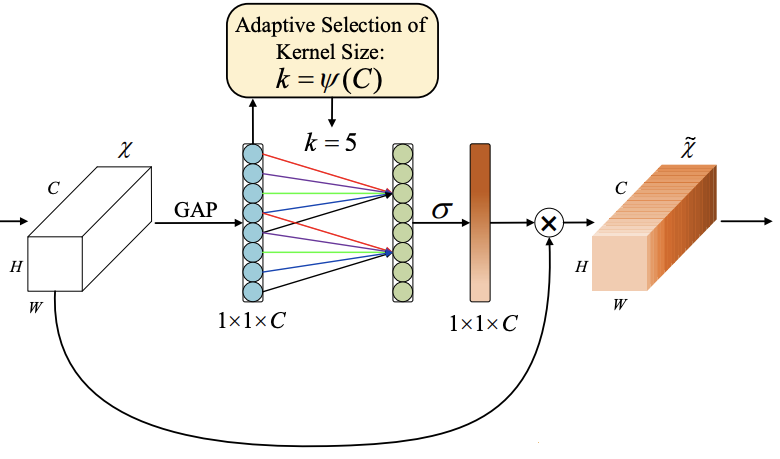} \\
    \small (g) ECA-Net \cite{wang2020eca}
  \end{tabular}
  \begin{tabular}[b]{c}
    \includegraphics[width=.2\paperwidth, height=.16\paperheight]{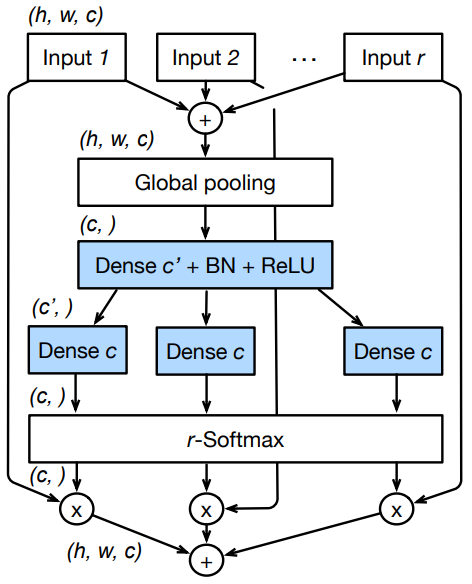} \\
    \small (h) RESNest \cite{zhang2020resnest}
  \tabularnewline
    \includegraphics[width=.2\paperwidth, height=.14\paperheight]{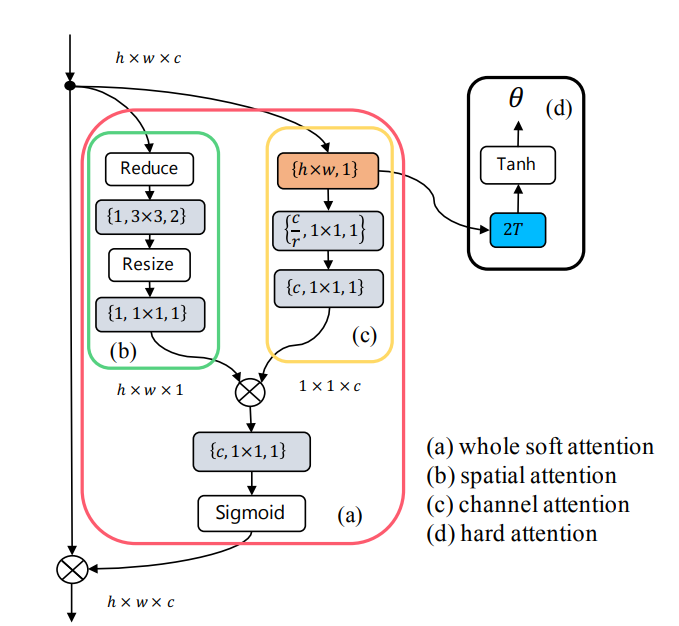} \\
    \small (i) Harmonious \cite{harmonious}
  \end{tabular}
  \caption{Core structures of the channel-based attention methods. Different methods to generate the attention scores, including squeeze and excitation~\cite{se}, splitting and squeezing~\cite{zhang2020resnest}, calculating the second order~\cite{fu2019dual} or efficient squeezing and excitation~\cite{wang2020eca}. Images are taken from the original papers and are best viewed in color.}
  \label{fig:channels}
\end{figure}

\subsubsection{Channel Attention}
\label{sec:channel}

\vspace{1.5mm}
\noindent   
~\\\textbf{Squeeze \& Excitation Attention}: 
The Squeeze-and-Excitation (SE) Block~\cite{hu2018squeeze}, shown in Figure~\ref{fig:channels}(a), is a unit designed to perform dynamic channel-wise feature attention. The SE attention takes the output of a convolution block and converts each channel to a single value via global average pooling; this process is called \enquote{squeeze}. The output channel ratio is reduced after passing through the fully connected layer and ReLU for adding non-linearity. The features are passed through the fully connected layer, followed by a sigmoid function to achieve a smooth gating operation. The convolutional block feature maps are weighted based on the side network's output, called the \enquote{excitation}. The process can be summarized as
\begin{equation}
f_s = \sigma( FC (ReLU( FC(f_g)) )),
\label{eq:SE_att}
\end{equation}
where $FC$ is the fully connected layer, $f_g$ is the average global pooling, $\sigma$ is the sigmoid operation. The main intuition is to choose the best representation of each channel in order to generate attention scores.

\vspace{1.5mm} 
\noindent   
\textbf{Efficient Channel Attention (ECA)~\cite{wang2020eca}} is based on squeeze $\&$ excitation network~\cite{hu2018squeeze} and aims to increase efficiency as well as decrease model complexity by removing the dimensionality reduction. ECA (see Fig~\ref{fig:channels}(g)) achieves cross-channel interaction locally by analyzing each channel and its $k$ neighbors, following channel-wise global average pooling but with no dimensionality reduction. ECA accomplishes efficient processing via fast 1D convolutions. The size $k$ represents the number of neighbors that can participate in one channel attention prediction, \ie the coverage of local cross-channel interaction.

\vspace{1.5mm} 
\noindent   
\textbf{Split-Attention Networks}: ResNest~\cite{zhang2020resnest}, a variant of ResNet~\cite{resnet}, uses split attention blocks as shown in Figure~\ref{fig:channels}(h). Attention is obtained by summing the inputs from previous modules and applying global pooling, passing through a composite function, \ie convolutional layer-batch normalization-ReLU activation. The output is again passed through convolutional layers. Afterward, a softmax is applied to normalize and multiply the values with the corresponding inputs. Finally, all the features are summed together. This mechanism is similar to the squeeze $\&$ excitation attention~\cite{hu2018squeeze}. ResNest is also a special type of squeeze $\&$ excitation where it squeezes the channels using average pooling and summing of the split channels.

\vspace{1.5mm} 
\noindent   
\textbf{Channel Attention in CBAM}: Convolutional Block Attention Module (CBAM)~\cite{woo2018cbam} employs channel attention and exploits the inter-channel feature relationship as each feature map channel is considered a feature detector focusing on the \enquote{what} part of the input image. The input feature map's spatial dimensions are squeezed to compute the channel attention followed by aggregation while using average-pooling and max-pooling to obtain two descriptors. These descriptors are forwarded to a three-layer shared multi-layer perceptron (MLP) to generate the attention map. Subsequently, the output of each MLP is summed element-wise and passed through a sigmoid function as shown in Figure~\ref{fig:channels}(b). In summary, the channel attention is computed as
\begin{equation}
f_{ch} = \sigma(  MLP(MaxPool(f)) + MLP(AvgPool(f))),
\label{eq:ch-atten} 
\end{equation}
where $\sigma$ denotes the sigmoid function, and $f$ represents the input features. The ReLU activation function is employed in MLP after each convolutional layer. If only average pooling is used, channel attention in CBAM is the same as Squeeze and Excitation (SE) attention~\cite{se}. 

\vspace{1.5mm} 
\noindent   
\textbf{Second-order Attention Network}:
For single image super-resolution, in~\cite{Dai_2019_CVPR}, the authors presented a second-order channel attention module, abbreviated as SOCA, to learn feature interdependencies via second-order feature statistics. A covariance matrix ($\Sigma$) is first computed and normalized using the features map from the previous network layers to obtain discriminative representations. The symmetric positive semi-definite covariance matrix is decomposed into $\Sigma = U\Lambda U^T$, where $U$ is orthogonal, and $\Lambda$ is the diagonal matrix with non-increasing eigenvalues. The power of eigenvalues $\Sigma = U\Lambda^\alpha U^T$ helps achieve the attention mechanism; if $\alpha < 1$, then the eigenvalues are larger than 1.0 will nonlinearly shrink while stretching others. The authors chose $\alpha < \frac{1}{2}$ based on previous work~\cite{li2017second}.  The subsequent attention mechanism is similar to SE~\cite{hu2018squeeze} as shown in Figure~\ref{fig:channels}(c), but instead of providing first-order statistics (\ie, global average pooling), the authors furnished second-order statistics (\ie, global covariance pooling).

\vspace{1.5mm} 
\noindent   
\textbf{High-Order Attention}: 
To encode global information and contextual representations, \cite{ding2020high}, Ding~\etal proposed High-order Attention (HA) with adaptive receptive fields and dynamic weights.  
HA mainly constructs a feature map for each pixel, including the relationships to other pixels. HA is required to address the issue of fixed-shape receptive fields that cause false prediction in the case of high-shape objects, \ie, similar shape objects. Specifically, graph transduction is used to form the final feature map after calculating the attention maps for each pixel. This feature representation updates each pixel position using the weighted sum of contextual information. High-order attention maps are calculated using Hadamard product \cite{horn1990hadamard, kim2016hadamard}. It is classified as channel attention because it generates attention scores from channels as in SE \cite{se}.

\vspace{1.5mm} 
\noindent   
\textbf{Harmonious attention}: proposes a joint attention module of soft pixel and hard regional attentions \cite{harmonious}. The main idea is to tackle the limitation of the previous attention modules in person Re-Identification by learning attention selection and feature representation jointly and solving the misalignment calibration caused by constrained attention mechanisms \cite{harmony1, harmony2, harmony3, harmony4}. Specifically, harmonious attention learns two types of soft attention (spatial and channel) in one branch and hard attention in the other. Moreover, it proposes cross-interaction attention harmonizing between these two attention types as shown in Figure~\ref{fig:channels}(i).

\input{table}

\vspace{1.5mm} 
\noindent   
\textbf{Auto Learning Attention}: Ma~\etal \cite{NEURIPS2020_103303dd} introduced a novel idea for designing attention automatically. The module, named Higher-Order Group Attention (HOGA), is in the form of a Directed Acyclic Graph (DAG) \cite{pham2018efficient, dag1, dag2, dag3} where each node represents a group and each edge represents a heterogeneous attention operation. There is a sequential connection between the nodes to represent hybrids of attention operations. Thus, these connections can be represented as K-order attention modules, where K is the number of attention operations. DARTS \cite{liu2018darts} is customized to facilitate the search process efficiently. This auto-learning module can be integrated into legacy architectures and performs better than manual ones. However, the core idea of attention modules remains the same as the previous architectures, \ie SE \cite{se}, CBAM \cite{woo2018cbam}, splat \cite{zhang2020resnest}, mixed \cite{chen2019mixed}.

\vspace{1.5mm} 
\noindent   
\textbf{Double Attention Networks}: Chen~\etal~\cite{chen20182} proposed a Double Attention Network (A2-Nets), which attends over the input image in two steps. The first step gathers the required features using bilinear pooling to encode the second-order relationships between entities, and the second step distributes the features over the various locations adaptively. In this architecture, the second-order statistics, mostly lost with other functions such as average pooling of SE~\cite{se}, of the pooled features are captured first by bilinear pooling. The attention scores are then calculated not from the whole image, such as \cite{m_nonlocal}, but from a compact bag, enriching the objects with the required context only. The first step \ie, feature gathering, uses the outer product $\sum_{\forall i} a_i b_i^T$ then softmax is used for attending the discriminative features. The second step \ie, distribution, is based on complementing each location with the required features where their summation is $1$. The complete design of A2-Nets is shown in Figure~\ref{fig:channels}(d). Experimental comparisons demonstrated that A2-Net improves performance better than SE and non-local networks and is more efficient regarding memory and time.

\vspace{1.5mm} 
\noindent   
\textbf{Dual Attention Network}: Jun~\etal~\cite{fu2019dual} presented a dual attention network for scene segmentation composed of position attention and channel attention working in parallel. The position attention aims to encode the contextual features in local ones. The attention process is straightforward: the input features $f_A$ are passed through three convolutional layers to generate three feature maps ($f_B$, $f_C$, and $f_D$), which are reshaped. Matrix multiplication is performed between the $f_B$ and the transpose of $f_C$, followed by softmax to obtain the spatial attention map. Again, matrix multiplication is performed between the generated $f_D$ features and the spatial attention map. Finally, the output is multiplied with a scalar and summed element-wise with the input features $f_A$ as shown in Figure~\ref{fig:channels}(f).

Although channel attention involves similar steps as position attention, it differs because the features are used directly without passing through convolutional layers. The input features $f_A$ are reshaped, transposed, multiplied (\ie, $f_A \times f_A'$), and then passed through the softmax layer to obtain the channel attention map. Moreover, the input features are multiplied with the channel attention map, followed by the element-wise summation, to give the final output as shown in Figure~\ref{fig:channels}(e).

\vspace{1.5mm} 
\noindent   
\textbf{Frequency Channel Attention}: Channel attention requires global average pooling as a pre-processing step. Qin~\etal~\cite{qin2020fcanet}  argued that the global average pooling operation could be replaced with frequency components. The frequency attention views the discrete cosine transform as the weighted input sum with the cosine parts. As global average pooling is a particular case of frequency-domain feature decomposition, the authors use various frequency components of 2D discrete cosine transform, including the zero-frequency component, \ie global average pooling.

\subsubsection{Spatial Attention}
\label{sec:spatial}
Unlike channel attention, which mainly generates channel-wise attention scores, spatial attention focuses on generating attention scores from spatial patches of the feature maps rather than the channels. However, the sequence of operations to generate attention is similar.

\begin{figure*}
  \centering
  \begin{tabular}[b]{c}
    \includegraphics[width=.3\paperwidth]{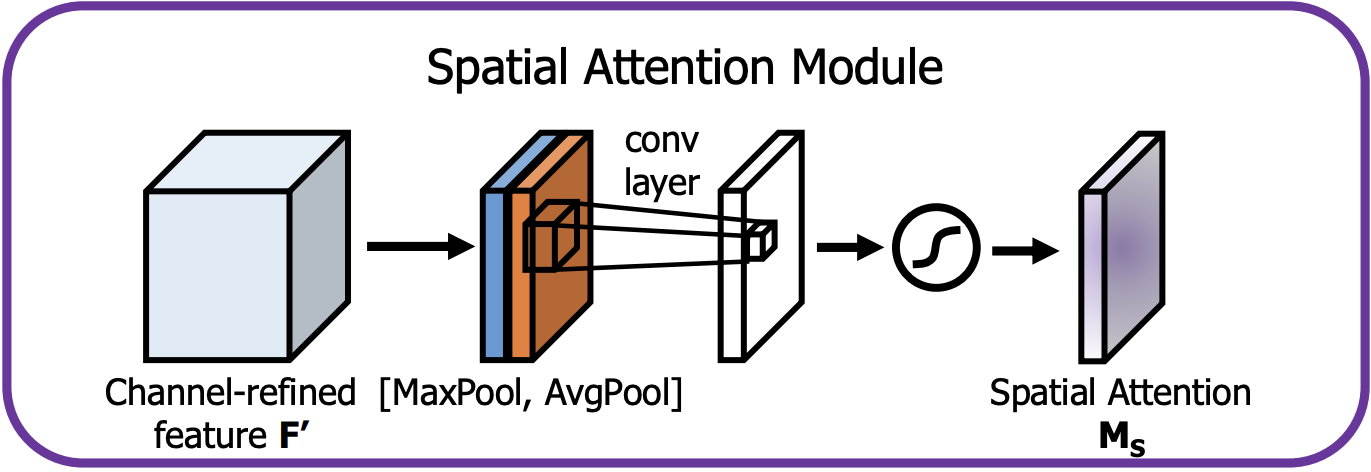}\\
    \small (a) Spatial Attention~\cite{woo2018cbam}
    \tabularnewline  
\includegraphics[width=.29\paperwidth]{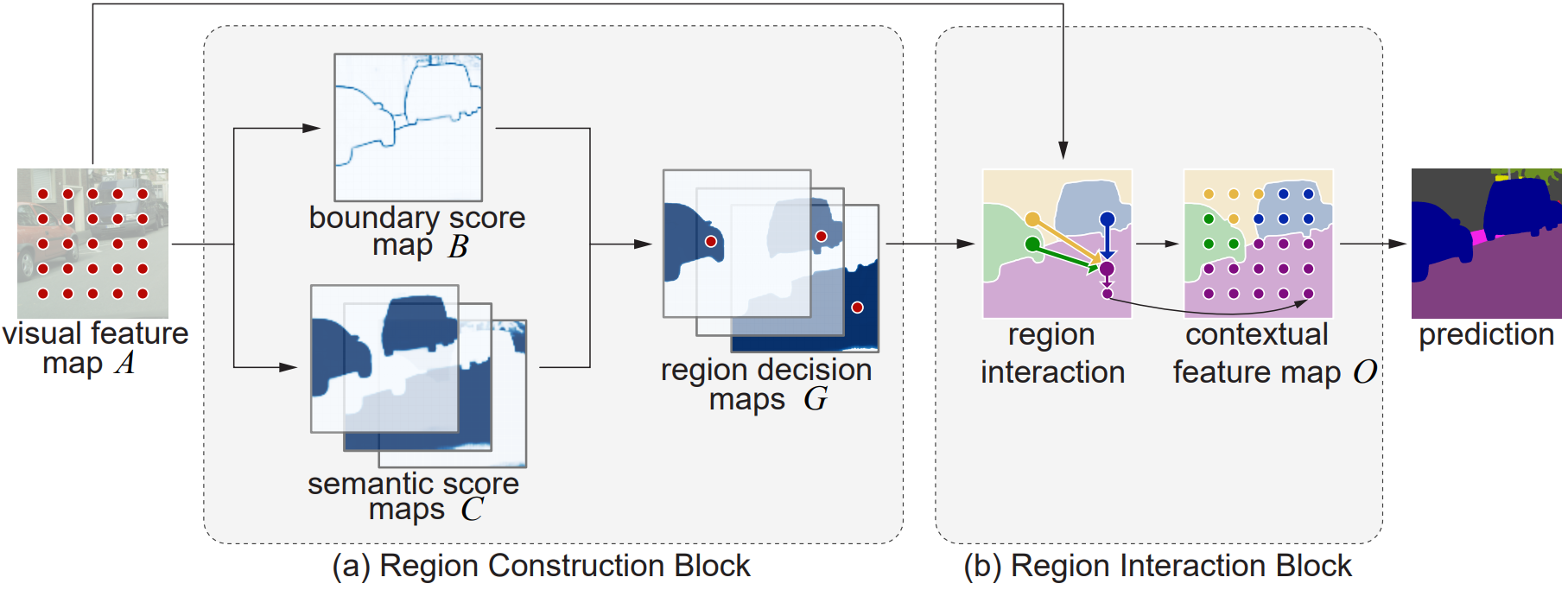} 
    \small \\(b) RANet~\cite{shen2020ranet}
    \tabularnewline
    \end{tabular}
\begin{tabular}[b]{c}
    \includegraphics[width=.32\paperwidth]{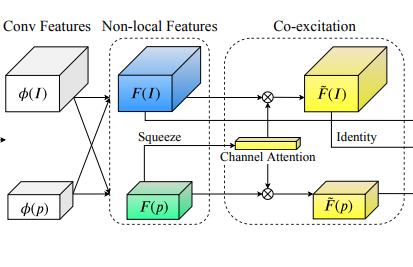} 
    \small \\(c) Co-excite \cite{NEURIPS2019_92af93f7}
  \end{tabular}
   \caption{The structures of the spatial-based attention methods, including RANet~\cite{shen2020ranet}, and Co-excite \cite{NEURIPS2019_92af93f7}. These methods focus on attending to the most important parts of the spatial map. The images are taken from~\cite{woo2018cbam,shen2020ranet,NEURIPS2019_92af93f7}.}
   \label{fig:spatial}
\end{figure*}

\noindent   
\textbf{Spatial Attention in CBAM} uses the inter-spatial feature relationships 
to complement the channel attention~\cite{woo2018cbam}. The spatial attention focuses on an informative part and is computed by applying average pooling and max pooling 
channel-wise, followed by concatenating both to obtain a single feature descriptor. Furthermore, a convolution layer on the concatenated feature descriptor is applied to generate a spatial attention map that encodes to emphasize or suppress. 
The feature map channel information is aggregated via average-pooled and max-pooled features and then concatenated and convolved to generate a 2D spatial attention map. The overall process is shown in Figure~\ref{fig:spatial}(a) and computed as
\begin{equation}
f_{sp} = \sigma( Conv_{7\times 7}([MaxPool(f); AvgPool(f)])),
\label{eq:sp-atten} 
\end{equation}
where $Conv_{7\times 7}$ denotes a convolution operation with the 7 $\times$ 7 kernel size and $\sigma$ represents the sigmoid function.

\vspace{1.5mm} 
\noindent   
\textbf{Co-attention \& Co-excitation}: Hsieh~\etal~\cite{NEURIPS2019_92af93f7} proposed co-attention and co-excitation to detect all the instances that belong to the same target for one-shot detection.  
The main idea is to enrich the extracted feature representation using non-local networks that encode long-range dependencies and second-order interactions \cite{m_nonlocal}. Co-excitation is based on squeeze-and-excite network \cite{se} as shown in Figure~\ref{fig:spatial}(e). In comparison, squeeze uses global average pooling~\cite{lin2013network} to reweight the spatial positions and co-excite bridges the feature of query and target. Encoding high-contextual representations using co-attention and co-excitation improves one-shot detector performance, achieving state-of-the-art results.

\vspace{1.5mm} 
\noindent   
\textbf{Spatial Pyramid Attention Network} abbreviated as SPAN~\cite{hu2020span}, 
was proposed for localizing multiple types of image manipulations. It is composed of three main blocks, \ie, feature extraction (head) module, pyramid spatial attention module, and decision (tail) module. The head module employs the Wider \& Deeper VGG Network as the backbone, while Bayer and SRM  layers extract features from visual artifacts and noise patterns. The spatial relationship of the pixels is achieved through five local self-attention blocks applied recursively, and the input of the self-attention is added to the output of the block to preserve the details as shown in Figure~\ref{fig:spatial}(c). After employing a sigmoid activation, these features are fed into the final tail module of 2D convolutional blocks to generate the output mask. 

\vspace{1.5mm} 
\noindent   
\textbf{Spatial-Spectral Self-Attention}: Figure~\ref{fig:spatial}(d) shows the architecture of spatial-spectral self-attention, which is composed of two attention modules, namely, spatial attention and spectral attention, both utilizing self-attention.

\begin{enumerate}
\item {Spatial Attention:}  To model the non-local region information, Meng~\etal~\cite{meng2020end} utilize a 3$\times$3 kernel to fuse the input features indicating the region-based correlation followed by a convolutional network mapping the fused features into Q $\&$ K. The kernel number indicates the heads' number and the size denotes the dimension. Moreover, the dimension-specified features from Q $\&$ K build the related attention maps and then modulate the corresponding dimension in a sequence to achieve the order-independent property. Finally, the features are forwarded to a deconvolution layer to finish the spatial correlation modeling.

\item{Spectral Attention}: First, the spectral channel samples are convolved with one kernel and flattened into  
a single dimension, set as the feature vector for that channel. The input feature is converted to Q $\&$ K, building attention maps for the spectral axis. The adjacent channels correlate more due to the image patterns on the exact location, denoted via a spectral smoothness on the attention maps. The similarity is indicated by normalized cosine distance as spectral embedding where each similar score is scaled and summed with the coefficients in the attention maps, which is then to modulate \enquote{Value} in self-attention, inducing spectral smoothness constraint.
\end{enumerate}

\vspace{1.5mm} 
\noindent   
\textbf{Pixel-wise Contextual Attention}: 
(PiCANet)~\cite{liu2018picanet} aims to learn accurate saliency detection. PiCANet generates a map at each pixel over the context region and constructs an accompanying contextual feature to enhance the feature representability at the local and global levels. To generate global attention, each pixel needs to \enquote{see} via ReNet~\cite{visin2015renet} with four recurrent neural networks sweeping horizontally and vertically. The contexts from directions, using biLSTM, are blended, propagating information from each pixel to all other pixels. Next, a convolutional layer transforms the feature maps to different channels, further normalized by a softmax function that weights the feature maps. The local attention is performed on a local neighborhood, forming a local feature cube where each pixel needs to \enquote{see} every other pixel in the local area using a few convolutional layers having the same receptive field as the patch. The features are then transformed to channel and normalized using softmax, which are further weighted and summed to get the final attention. 

\vspace{1.5mm} 
\noindent   
\textbf{Pyramid Feature Attention} extracts features from different levels of VGG~\cite{zhao2019pyramid}. The low-level features extracted from lower layers of VGG are provided to the spatial attention mechanism~\cite{woo2018cbam}, and the high-level features obtained from the higher layers are supplied to a channel attention mechanism~\cite{woo2018cbam}. The term feature pyramid attention originates from the VGG features obtained from different layers.

\vspace{1.5mm} 
\noindent   
\textbf{Spatial Attention Pyramid}: For unsupervised domain adaptation, Li~\etal~\cite{li2020spatial} introduced a spatial attention pyramid that takes features from multiple average pooling layers with various sizes operating on feature maps. These features are forwarded to spatial attention followed by channel-wise attention. All the features after attention are concatenated to form a single semantic vector.

\vspace{1.5mm} 
\noindent   
\textbf{Region Attention Network}: 
(RANet)~\cite{shen2020ranet} was proposed for semantic segmentation. 
It consists of novel network components, the Region Construction Block (RCB) and the Region Interaction Block (RIB), for constructing the contextual representations as illustrated in Figure~\ref{fig:spatial}(b). The RCB analyzes the boundary and semantic score maps jointly to compute the attention region score for each image pixel pair. A high attention score indicates that the pixels are from the same object region, dividing the image into various object regions. Subsequently, the RIB takes the region maps and selects the representative pixels in different areas where each representative pixel receives the context from other pixels to effectively represent the object region's local content. Furthermore, capturing the spatial and category relationship between various objects communicating the representative pixels in the different regions yields the global contextual representation to augment the pixels, eventually forming the contextual feature map for segmentation.

\subsubsection{Self-attention } 
\label{sec:self}
Self-attention, also known as \emph{intra-attention}, is an attention mechanism that encodes the relationships between all the input entities. It is a process that enables input sequences to interact with each other and aggregate the attention scores, which illustrate how similar they are. The main idea is to replicate the feature maps into three copies and then measure their similarity. Apart from the channel- and spatial-wise attention that uses the physical feature maps, self-attention replicates feature copies to measure long-range dependencies. However, self-attention methods use channels to calculate attention scores. Cheng~\etal extracted the correlations between the words of a single sentence using Long-Short-Term Memory (LSTM) \cite{m_self_attention}. An attention vector is produced from each hidden state during the recurrent iteration, which attends all the responses in a sequence for this position. In \cite{m_self_attention_parikah}, a decomposable solution was proposed to divide the input into sub-problems, which improved the processing efficiency compared to \cite{m_self_attention}. The attention vector is calculated as an alignment factor to the content (bag-of-words). Although these methods introduced the idea of self-attention, they are very expensive in terms of resources and do not consider contextual information. Also, RNN models process input sequentially; hence, it is difficult to parallelize or process large-scale schema efficiently.

\begin{figure}[tbp]
\centering
\begin{tabular}{c@{}c@{}c@{}}
\includegraphics[width=.3\textwidth]{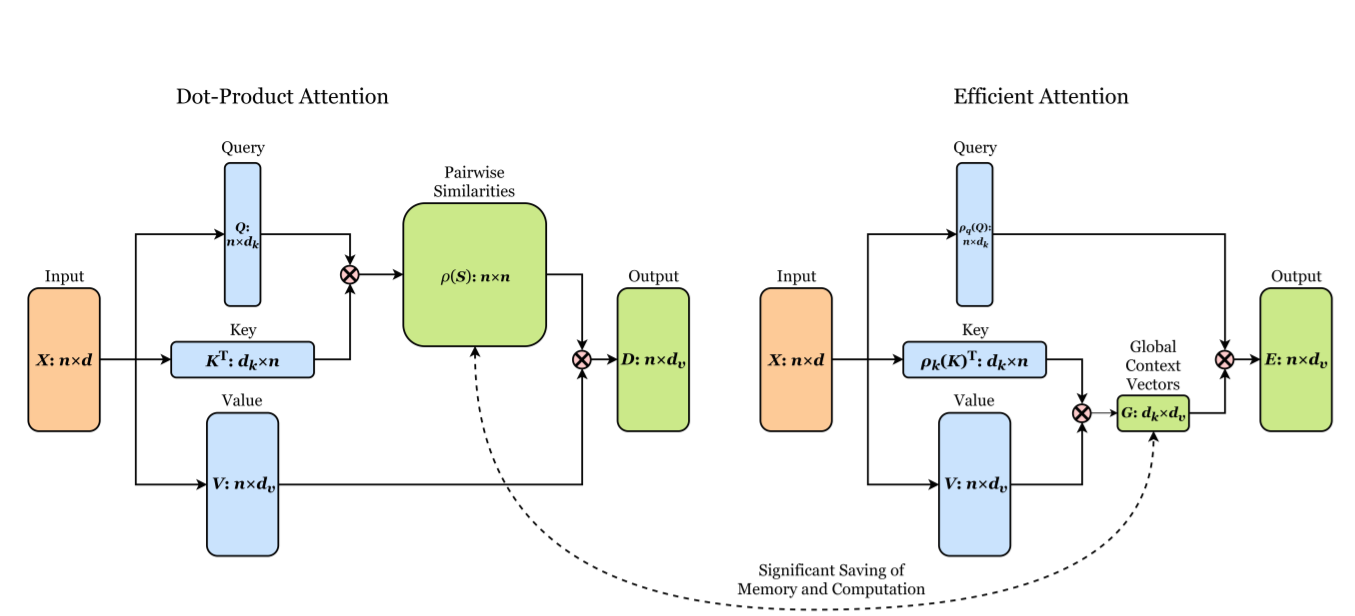} &
\includegraphics[width=.2\textwidth]{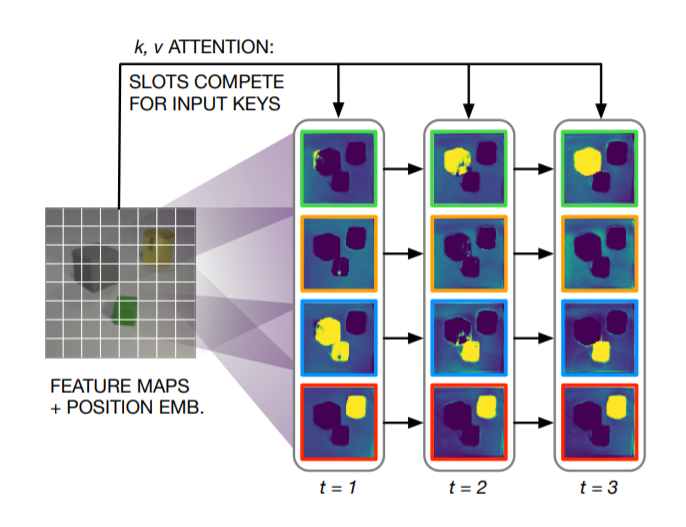} &
\includegraphics[width=.35\textwidth]{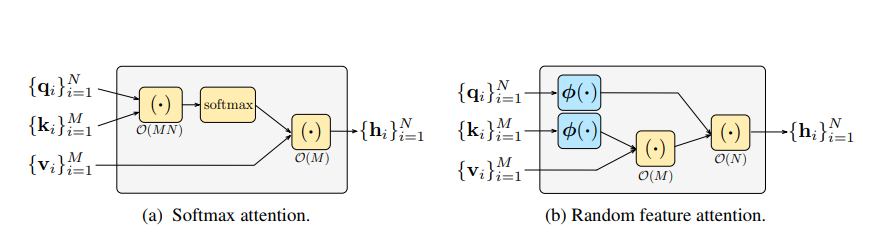} \\
    \small (a) Efficient Attention~\cite{Efficient_attention}&
    \small (b) Slot Attention~\cite{slot}&
    \small (c) RFA~\cite{peng2021random}\\
\includegraphics[width=.33\textwidth]{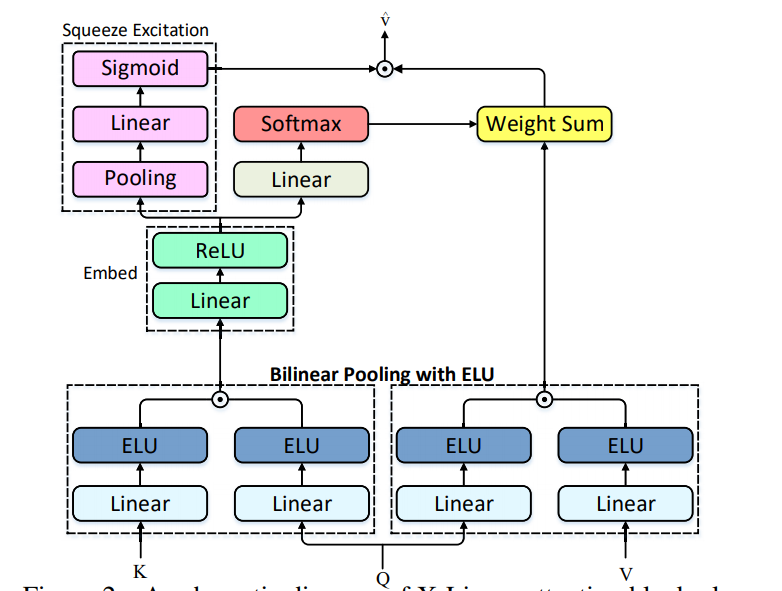} &
\includegraphics[width=.33\textwidth]{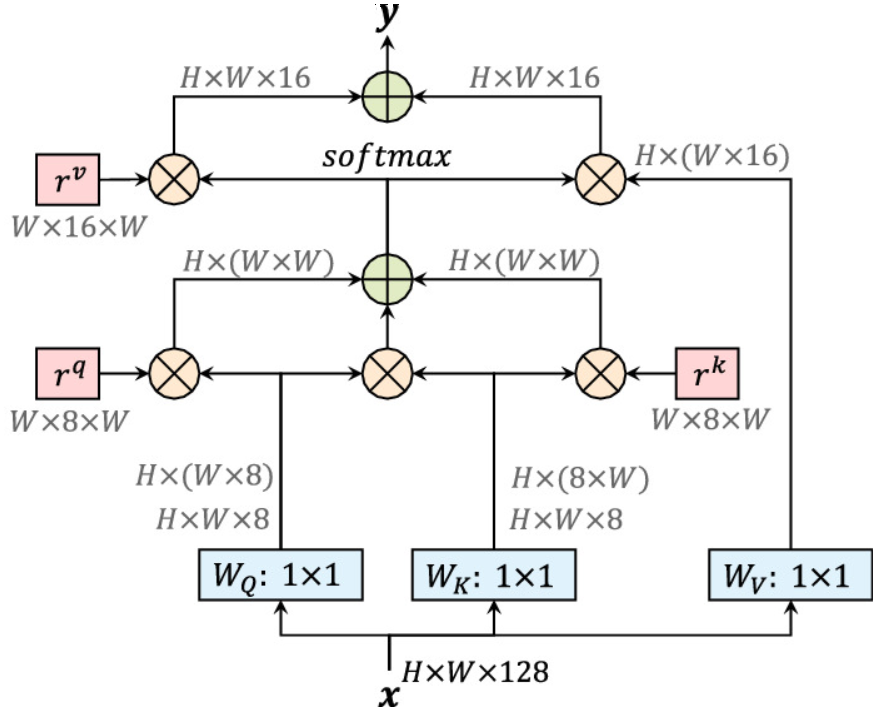} &
\includegraphics[width=.24\textwidth]{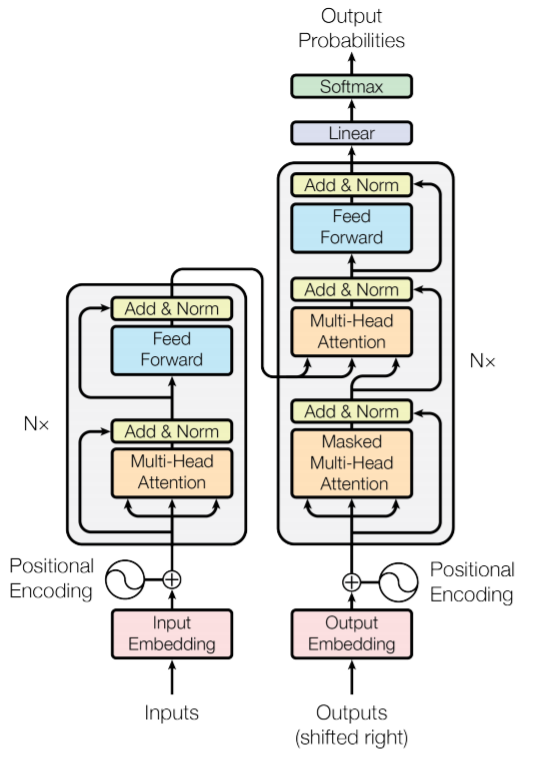}\\
    \small (d) X-Linear~\cite{pan2020x}&
    \small (e) Axial~\cite{axial_attention}&
    \small (f) Transformer~\cite{m_transformers}\\
\end{tabular} 
\caption{ The architectures of self-attention methods: Transformers~\cite{m_transformers}, Axial attention~\cite{axial_attention}, X-Linear~\cite{pan2020x}, Slot~\cite{slot} and RFA~\cite{peng2021random} (pictures taken from the corresponding articles). These methods are self-attention, which generates the scores by measuring the similarity between two maps of the same input. However, there is a difference in the way of processing.}
\label{fig:self_attentions}
\end{figure}

\vspace{1.5mm} 
\noindent   
\textbf{Transformers}: Vaswani~\etal~\cite{m_transformers} proposed a new method, called transformers, based on the self-attention concept without convolution or recurrent modules. As shown in Figure~\ref{fig:self_attentions} (f), it is mainly composed of encoder-decoder layers, where the encoder comprises a self-attention module followed by a positional feed-forward layer, and the decoder is the same as the encoder except that it has an encoder-decoder attention layer in between. Positional encoding is represented by a sine wave that incorporates the passage of time as input before the linear layer. This positional encoding serves as a generalization term to help recognize unseen sequences and encode relative positions rather than absolute representations. Algorithm~\ref{algorithm:self-attention} shows the detailed steps of calculating self-attention (multi-head attention) using transformers. Although transformers have achieved much progress in the text-based models, they lack in encoding the sentence context because they calculate the word's attention for the left-side sequences. To address this issue, Bidirectional Encoder Representations from
Transformers (BERT) learn the contextual information by encoding both sides of the sentence jointly \cite{m_BERT}.
 
\begin{algorithm}[t]
\SetAlgoLined
\SetKwInOut{Input}{Input}
\SetKwInOut{Output}{Output}
 \Input {set of sequences $(x_1, x_2, \cdots, x_n)$ of an entity $\mathbf{X} \in \mathbf{R}$}
 \Output{attention scores of $\mathbf{X}$ sequences.}
 {
Initialize weights: Key ($\mathbf{W_K}$), Query ($\mathbf{W_Q}$), Value ($\mathbf{W_V}$) for each input sequence. \\
Derive Key, Query, Value  for each input sequence and its corresponding weight, such that $\mathbf{Q = XW_Q}$,  $\mathbf{K = XW_K}$,  $\mathbf{V = XV_Q}$, respectively.\\
Compute attention scores by calculating the dot product between the query and key.\\
Compute  the scaled-dot product attention for these scores and Values $\mathbf{V}$,  
    \[ \mathrm{softmax}  \left( \frac{\mathbf{QK^T}}{\sqrt{d_k}}\right)\mathbf{V}.\]\\
repeat steps from 1 to 4  for all the heads \\
}
 \caption{The main steps of generating self-attention by transformers (multi-head attention)}
 \label{algorithm:self-attention}
\end{algorithm}


\vspace{1.5mm} 
\noindent   
\textbf{Swin Transformers}: 
Traditional Transformers were first introduced in the text processing field, and ViT~\cite{image_transfomer} proposed it for visual tasks. However, it was unsuitable for dense processing tasks such as detection and segmentation that need fine-grain processing. Swin Transformers~\cite{liu2021swin} tackled this issue by processing self-attention on a small image area. The critical element of Swin Transformers is shift tokens that shift the partitions between attention modules. It has a hierarchical design allows the processing to go deeper into the feature maps. It achieved state-of-the-art in the visual tasks as it became the main module in the backbone architectures.  

\vspace{1.5mm} 
\noindent   
\textbf{Deformable Attention Transformers}: 
 Dense Transformers~\cite{image_transfomer} has progressed the entire visual paradigms; however, they suffer from high memory usage, computational cost and convergence time. Sparse Transformers~\cite{liu2021swin} tried to tackle the drawbacks of dense transformers by attending to small areas rather than the whole image. However, the long relationships for the entire image are lost, affecting the image's global relationships. Deformable Attention Transformers (DAT) apply the same methodology as sparse transformers in small areas but use deformable attention to select the most relevant areas. In this way, DAT successfully controls the attention range by combining the strength points of dense and sparse transformers. It achieved state-of-the-art computer vision tasks such as detection and segmentation.

\vspace{1.5mm} 
\noindent   
\textbf{Standalone self-attention}: As stated above, convolutional features do not consider the global information due to their local-biased receptive fields. Instead of augmenting attentional features to the convolutional ones, Ramachandran~\etal~\cite{m_standalone} proposed a fully attentional network that replaces spatial convolutions with self-attentional modules. The convolutional stem (the first few convolutions) captures spatial information. They designed a small kernel (\eg $n \times n$) instead of processing the whole image simultaneously. This design built a computationally efficient model that enables processing images with their original sizes without downsampling. The computation complexity is reduced to $\mathcal{O}(hwn^2)$, where $h$ and $w$ denote height and width, respectively. A patch is extracted as a query along with each local patch, while the identity image is used as Value and Key. Calculating the attention maps follows the same steps as in Algorithm \ref{algorithm:self-attention}. Although standalone self-attention shows competitive results compared to convolutional models, it suffers from encoding positional information.

\vspace{1.5mm}  
\noindent  

\vspace{1.5mm}
\noindent   
\textbf{Clustered Attention}: To address the computational inefficiency of transformers, Vyas~\etal~\cite{m_clsutered} proposed a clustered attention mechanism that relies on the idea that correlated queries follow the same distribution around Euclidean centers. Based on this idea, they use the K-means algorithm with fixed centers to group similar queries. Instead of calculating the queries for attention, 
they are calculated for clusters' centers. Therefore, the total complexity is minimized to a linear form $\mathcal{O}(qc)$, where $q$ is the number of the queries while $c$ is the  
cluster number.  

\vspace{1.5mm} 
\noindent   
\textbf{Slot Attention}: Locatello~\etal~\cite{scouter_slot} proposed slot attention, an attention mechanism that learns the objects' representations in a scene. It generally learns to disentangle the image into a series of slots. As shown in Figure~\ref{fig:self_attentions}(d), each slot represents a single object. The slot attention module is applied to a learned representation $h \in \mathbb{R}^{W \times H \times D}$, where $H$ is the height, $W$ is the width, and $D$ is the representation size. SAM has two main steps: learning $n$ slots using an iterative attention mechanism and representing individual objects (slots). Inside each iteration, two operations are implemented: 1) slot competition using softmax followed by normalization according to slot dimension using this equation.

\begin{equation}
    a  = Softmax \bigg(  \frac{1}{\sqrt{D}} n (h). q(c)^T\bigg).
\end{equation}
2) An aggregation process for the attended representations with a weighted mean
\begin{equation}
    r  = Weightedmean \bigg(a, v(h)  \bigg),
\end{equation}
where $k, q, v$ are learnable variables as showed in \cite{m_transformers}. Then, a feed-forward layer is used to predict the slot representations $s=fc(r)$.

Slot attention is based on Transformer-like attention \cite{m_transformers} on top of CNN-feature extractors. Given an image $\mathbb{I}$, the slot attention parses the scene into a set of slots, each one referring to an object $(z, x, m)$, where $z$ is the object feature, $x$ is the input image and $m$ is the mask. In the decoders, convolutional networks are used to learn slot representations and object masks. The training process is guided by $\ell_2$-norm loss
\begin{equation}
\mathbb{L} = \bigg\rVert \bigg(   \sum_{k=1}^K mx_k\bigg) - \mathbb{I} \bigg\lVert_2^2
\end{equation}

Following the slot-attention module, Li~\etal developed an explainable classifier based on slot attentions \cite{scouter_slot}. This method aims to find positive and negative supports for a class $l$. In this way, a classifier can also be explained rather than being completely black-box.  
The primary entity of this work is xSlot, a variant of slot attention~\cite{slot}, which is related to a category and gives confidence for the inclusion of this category in the input image. 

\vspace{1.5mm} 
\noindent   
\textbf{Efficient Attention}: Using Asymmetric Clustering (SMYRF) Daras~\etal~\cite{SMYRF} proposed symmetric Locality Sensitive Hashing (LSH) clustering in a novel way to reduce the size of attention maps, therefore, developing efficient models. They observed that attention weights are sparse, as the attention matrix is low-rank. As a result, pre-trained models pertain to decay in their values. SMYRF addresses this issue by approximating attention maps through balanced clustering produced by asymmetric transformations and an adaptive scheme. SMYRF is a drop-in replacement for pre-trained models for normal dense attention. SMYRF showed significant effectiveness in memory, performance, and speed without retraining models after integrating this module. Therefore, the feature maps can be scaled up to include contextual information. In some models, the memory usage is reduced by $50\%$. Although SMYRF enhanced memory usage in self-attention models, the improvement over efficient attention models is marginal (see Figure~\ref{fig:self_attentions} (a)).

\vspace{1.5mm} 
\noindent   
\textbf{Random Feature Attention}: Transformers have a major shortcoming with regard to time and memory complexity, which hinders attention scaling up and thus limits higher-order interactions. Peng~\etal~\cite{peng2021random} proposed reducing the space and time complexity of transformers from quadratic to linear. They simply enhance softmax approximation using random functions. Random Feature Attention (RFA) \cite{rawat2019sampled} uses a variant of softmax that is sampled from simple distribution-based Fourier random features \cite{rahimi2007random, yang2014quasi}. Using the kernel trick $exp(x.y) \approx \phi(x).\phi(y)$ of \cite{hofmann2008kernel}, the softmax approximation is reduced to linear as shown in Figure~\ref{fig:self_attentions} (c). Moreover, the similarity of RFA connections and recurrent networks helps develop a gating mechanism to learn recency bias \cite{lstm, cho2014learning, schmidhuber1992learning}. RFA can be integrated into backbones easily to replace the normal softmax without increasing the number of parameters, only $0.1\%$ increase. Plugging RFA into a transformer shows comparable results to softmax while gating RFA outperformed it in language models. RFA executes 2$\times$ faster than a conventional transformer.

\vspace{1.5mm} 
\noindent   
\textbf{Non-local Networks}: 
Recent breakthroughs in artificial intelligence are mostly based on the success of Convolution Neural Networks (CNNs) \cite{m_deep_learning, resnet}. In particular, they can be processed in parallel mode and are inductive biases for the extracted features. However, CNNs fail to learn the context of the whole image due to their local-biased receptive fields. Therefore, long-range dependencies should be considered in CNNs. In \cite{m_nonlocal}, Wang~\etal proposed non-local networks to alleviate the bias of CNNs towards the local information and fuse global information into the network. It augments each pixel of the convolutional features with the contextual information, the weighted sum of the whole feature map. In this manner, the correlated patches in an image are encoded in a long-range fashion. Non-local networks showed significant improvement in long-range interaction tasks such as video classification \cite{m_kinect} as well as low-level image processing \cite{m_non_denoise, non_local_advers}. Non-local model attention in the network in a graphical fashion \cite{m_graph_atten}. However, stacking multiple non-local modules in the same stage shows instability and ill-pose in the training process~\cite{m_non_local_diffuse}. In~\cite{liu2020learning}, Liu~\etal uses non-local networks to form self-mutual attention between two modalities (RGB and Depth) to learn global contextual information. The idea is straightforward, \ie, to sum the corresponding features before softmax normalization such that $\mathrm{softmax}(f^r (\mathbb{X}^r)+\alpha^d \bigodot f^d (\mathbb{X}^d))$ for RGB attention and vice versa.

\vspace{1.5mm} 
\noindent   
\textbf{Non-Local Sparse Attention (NLSA)}: Mei~\etal \cite{mei2021image} proposed a sparse non-local network to combine the benefits of non-local modules to encode long-range dependencies and sparse representations for robustness. Deep features are split into different groups (buckets) with high inner correlations. Locality Sensitive Hashing (LSH) \cite{gionis1999similarity}  is used to find similar features to each bucket. Then, the Non-Local block processes the pixel within its bucket and similar ones. NLSA reduces the complexity to asymptotic linear from quadratic and uses the power of sparse representations to focus on informative regions only.

\vspace{1.5mm} 
\noindent   
\textbf{X-Linear Attention}: Bilinear pooling is a calculation process that computers the outer product between two entities rather than the inner product \cite{bilinear4, bilinear3, bilinear2, bilinear1} and has shown the ability to encode higher-order interaction and thus encourage more discriminability in the models. Moreover, it yields compact models with the required details even though it compresses the representations \cite{bilinear2}. In particular, bilinear applications have shown significant improvements in fine-grained visual recognition \cite{fine3, fine2, fine1} and visual question answering \cite{yu2017multi}. As Figure~\ref{fig:self_attentions}(e) depicts, a low-rank bilinear pooling is performed between queries and keys, and hence, the $2^{nd}$-order interactions between keys and queries are encoded. Through this query-key interaction, spatial-wise and channel-wise attention are aggregated with the values. The channel-wise attention is the same as squeeze-excitation attention \cite{se}. The final output of the x-linear module is aggregated with the low-rank bilinear of keys and values \cite{pan2020x}. They claimed that encoding higher interactions requires only repeating the x-linear module accordingly (\eg three iterative x-linear blocks for $4^{th}$-order interactions). Modeling infinity-order interaction is also explained using $h$ Exponential Linear Unit \cite{barron2017continuously}. X-Linear attention module proposes a novel mechanism, different from transformer~\cite{m_transformers}. It is able to encode the relations between input tokens without positional encoding with only linear complexity as opposed to quadratic in the transformer.

\vspace{1.5mm} 
\noindent   
\textbf{Axial-Attention}: Wang~\etal~\cite{axial_attention} proposed axial attention to encode global information and long-range context for the subject. Although conventional self-attention methods use fully-connected layers to encode non-local interactions, they are costly given their dense connections \cite{m_transformers,bert,detr,image_transfomer}. Axial uses self-attention models in a non-local way without any constraints. Simply put, it factorizes the 2D self-attentions in two axes (width and height) of 1D-self attentions. This way, axial attention shows effectiveness in attending over wide regions. Moreover, unlike \cite{bam, ramachandran2019stand, hu2019local}, axial attention uses positional information to include contextual information agnostic. Axial attention reduces the computational complexity to $\mathcal{O}(hwm)$. Also, axial attention showed competitive performance compared to full-attention models~\cite{attention_augmented,m_standalone}, and convolutional ones as well \cite{resnet, huang2017densely}.

\vspace{1.5mm} 
\noindent   
\textbf{Efficient Attention Mechanism}: Conventional attention mechanisms are built on double matrix multiplication that yields quadratic complexity $n \times n$ where $n$ is the size of the matrix. Many methods propose efficient architectures for attention \cite{kitaev2019reformer, Efficient_attention, wu2021centroid, kim2020fastformers}. In~\cite{Efficient_attention}, Zhuoran~\etal used the associative feature of matrix multiplication and suggested efficient attention. Formally, instead of using dot-product of the form $\rho (QK^T)V$, they process it in an efficient sequence $\rho_q(Q)(\rho_k(K)^T V)$ where $\rho$ denotes a normalization step. Regarding the normalization of $\mathrm{softmax}$, it is performed twice instead of once at the end. Hence, the complexity is reduced from quadratic $\mathcal{O}(n^2)$ to linear $\mathcal{O}(n)$. A simple change reduces the complexity of processing and memory usage to enable the integration of attention modules in large-scale tasks.

\subsubsection{Arithmetic Attention}
\label{sec:arithmetic}

This part introduces arithmetic attention methods such as dropout, mirror, reverse, inverse, and reciprocal. We named it arithmetic because these methods differ from the above techniques even though they use their core. However, these methods mainly produce the final attention scores from simple arithmetic equations such as the reciprocal of the attention, \etc

\begin{figure*}
  \centering
  \begin{tabular}{cc}
    \includegraphics[width=.5\paperwidth]{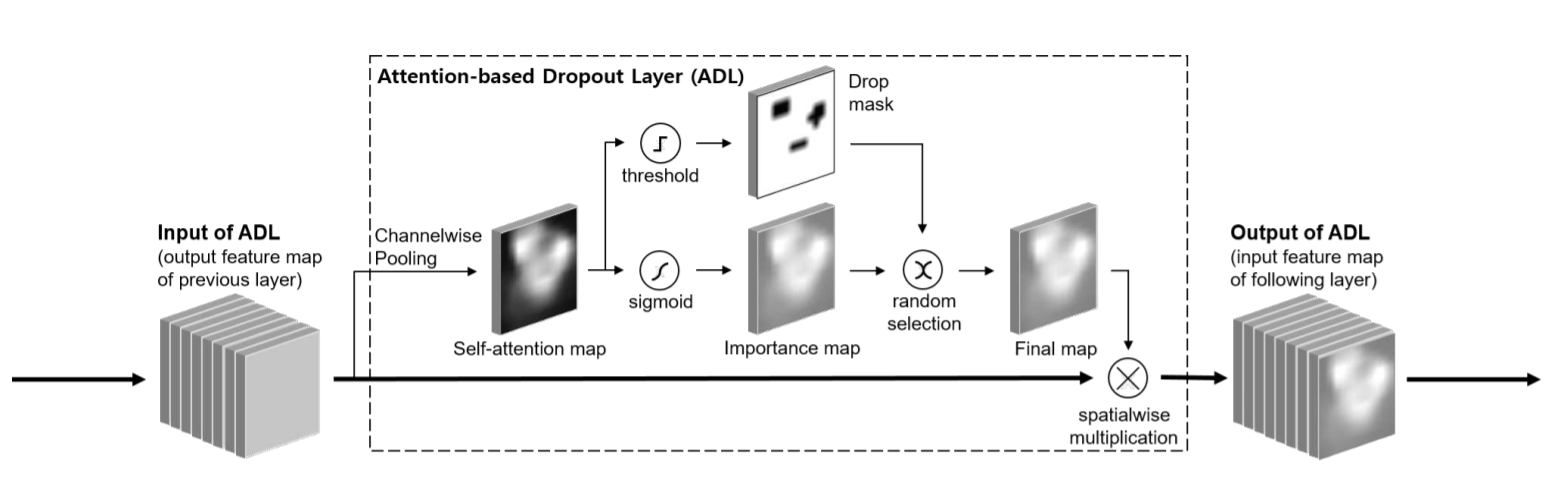} &   \includegraphics[width=.23\paperwidth]{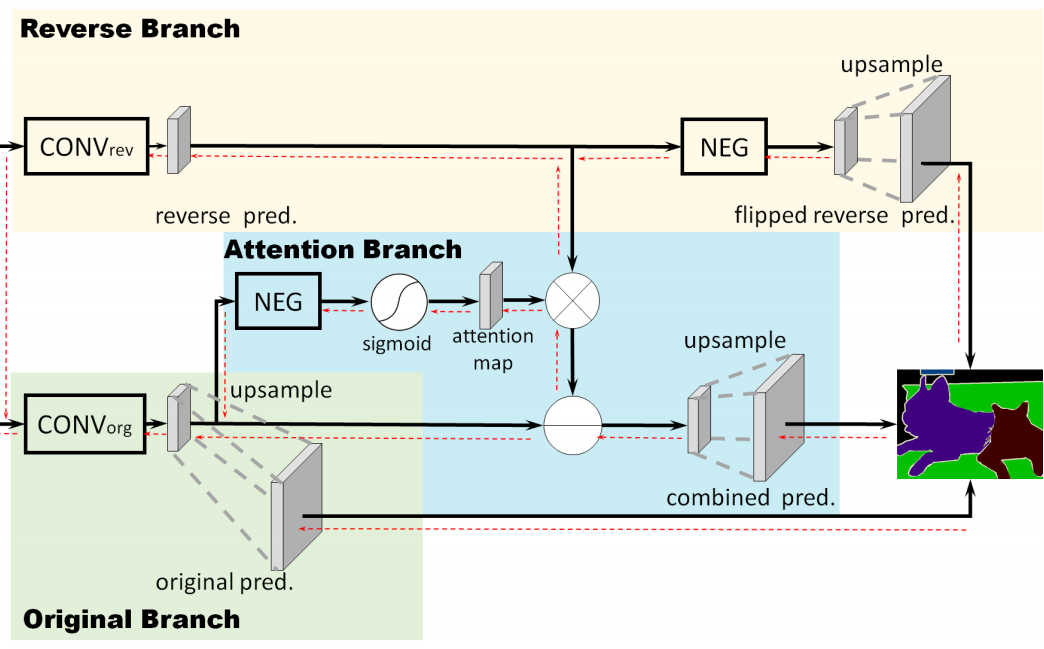}  \\
    \small (a) Attention-Based Dropout~\cite{choe2019attention}& 
    \small (b) Reverse Attention~\cite{BMVC2017_18}
  \end{tabular}
   \caption{The arithmetic-based attention methods i.e. Attention-based Dropout~\cite{choe2019attention} and  Reverse Attention  \cite{BMVC2017_18}. Images are taken from the original papers. These methods use arithmetic operations to generate the attention scores such as reverse, dropout or reciprocal. }
     \label{fig:arithmetic}
\end{figure*}

\vspace{1.5mm} 
\noindent   
\textbf{Attention-based Dropout Layer}: In weakly-supervised object localization, detecting the whole object without location annotation is a challenging task \cite{wsol1, wsol1, wsol2}. Choe~\etal \cite{choe2019attention} proposed using the dropout layer to improve the localization accuracy through two steps: making the whole object location even by hiding the most discriminative part and attending over the whole area to improve the recognition performance. As Figure~\ref{fig:arithmetic} (a) shows, ADL has two branches: 1) drop mask to conceal the discriminative part, which is performed by a threshold hyperparameter where values bigger than this threshold are set to zero and vice versa, and 2) importance map to give weight for the channels contributions by using a sigmoid function. Although the proposed idea is simple, experiments showed it is efficient (gained 15 $\%$ more than the state-of-the-art).

\vspace{1.5mm} 
\noindent   
\textbf{Mirror Attention}: In a line detection application \cite{jin2020semantic}, Lee~\etal developed mirrored attention to learn more semantic features. They flipped the feature map around the candidate line and then concatenated the feature maps together. In case the line is not aligned, zero padding is applied.

\vspace{1.5mm} 
\noindent   
\textbf{Reverse Attention}: Huang~\etal~\cite{BMVC2017_18} proposed the negative context (\eg what is not related to the class) in training to learn semantic features. They were motivated by less discriminability between the classes in the high-level semantic representations and the weak response to the correct class from the latent representations. The network is composed of two branches: the first one learns discriminative features using convolutions for the target class, and the second one learns the reverse attention scores that are not associated with the target class. These scores are aggregated together to form the final attentions shown in Figure~\ref{fig:arithmetic} (b). A deeper look inside the reverse attention shows that it is mainly dependent on negating the extracted features of convolutions followed by sigmoid $\mathrm{sigmoid} (-F_{conv})$. However, for convergence, this simple equation is changed to be $\mathrm{sigmoid}(\frac{1}{ReLU(F_{conv})+0.125} - 4)$. On semantic segmentation datasets, reverse attention achieved significant improvement over the state-of-the-art. In a similar work, Chen~\etal~\cite{chen2018reverse} proposed using reverse attention for salient object detection. The main intuition was to erase the final predictions of the network and, hence, learn the missing parts of the objects. However, the calculation method of attention scores is different from \cite{BMVC2017_18}, whereas they used  $1 - \mathrm{sigmoid}(F_{i+1})$ where $F_{i+1}$ denoted the features of the next stage. 

\subsubsection{Multi-modal attentions}
\label{sec:Multi-modal}

As the name reveals, multi-modal attention is proposed to handle multi-modal tasks, using different modalities to generate attention, such as text and image. It should be noted that some attention methods below, such as Perceiver~\cite{jaegle2021perceiver} and Criss-Cross \cite{context1, context2}, are transformer types~\cite{m_transformers}, but are customized for multi-modal tasks by including text, audio, and image.

\begin{figure}[t]
  \centering
  \begin{tabular}[b]{c}
    \includegraphics[width=.5\paperwidth]{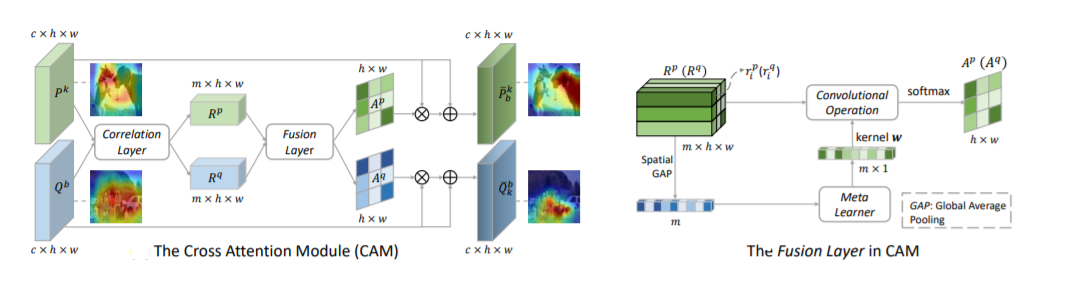} \\
    \small (a) Cross Attention~\cite{hou2019cross}
  \end{tabular} 
  \begin{tabular}[b]{c}
    \includegraphics[width=.25\paperwidth]{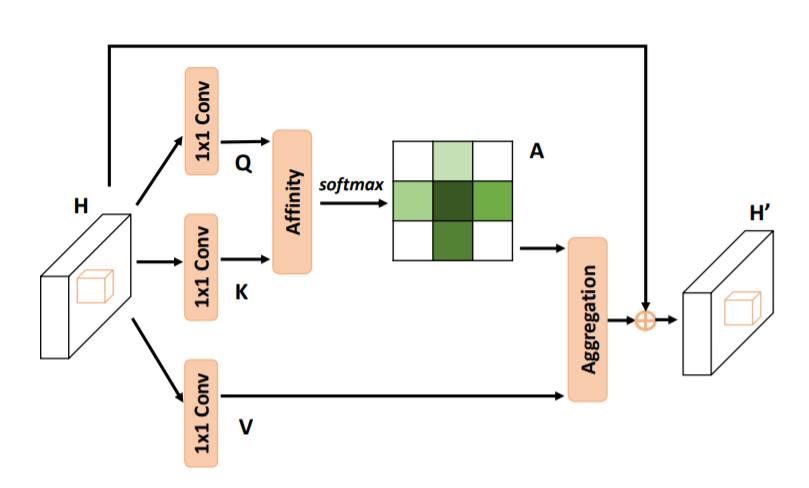} \\
    \small (b) Criss-Cross Attention~\cite{huang2019ccnet}
  \end{tabular} 
  \begin{tabular}[b]{c}
    \includegraphics[width=.48\paperwidth]{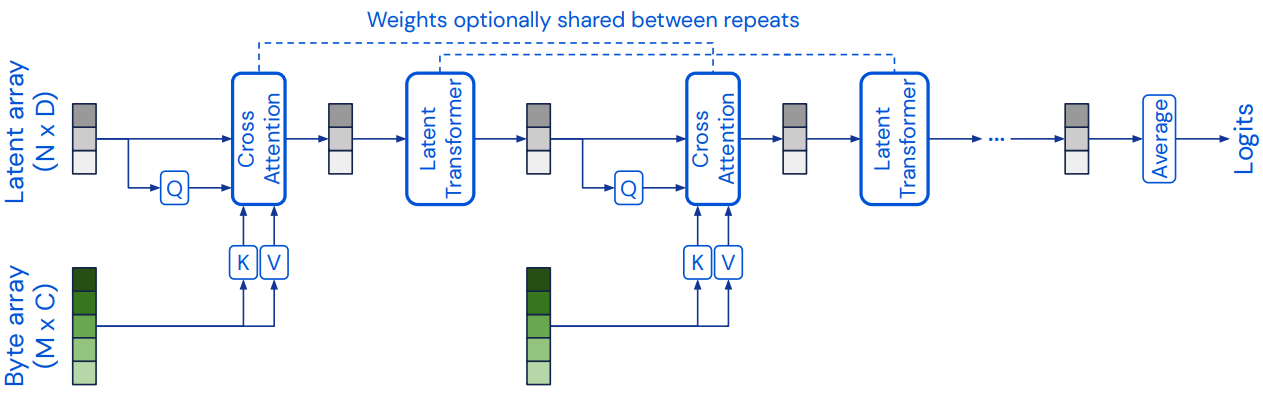} \\
    \small (c) Perceiver~\cite{jaegle2021perceiver}
  \end{tabular}
  \begin{tabular}[b]{c}
    \includegraphics[width=.28\paperwidth]{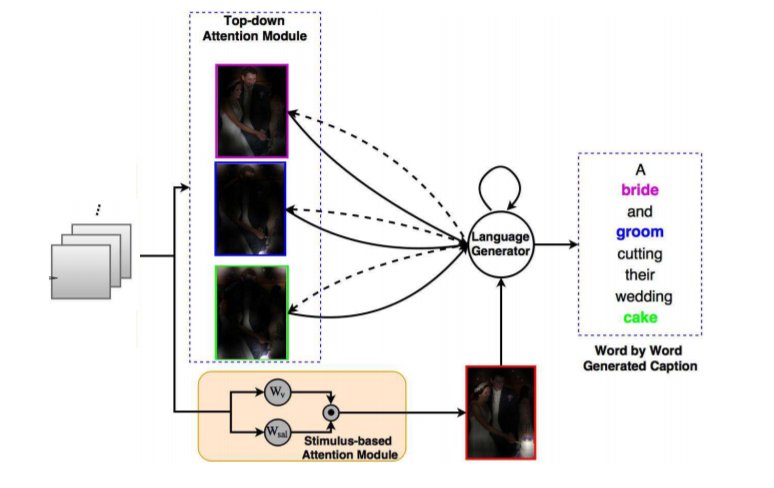} \\
    \small (d) Boosted Attention~\cite{chen2018boosted}
  \end{tabular}
   \caption{Multi-modal attention methods consisting of Attention-based Perceiver~\cite{jaegle2021perceiver}, Criss-Cross~\cite{hou2019cross}, Boosted attention~\cite{chen2018boosted}, Cross-attention module~\cite{hou2019cross}. The mentioned methods employ multi-modalities to generate attention scores. Images are taken from the original papers.}
   \label{fig:multimodals}
\end{figure}

\vspace{1.5mm} 
\noindent   
\textbf{Cross Attention Network}: In \cite{hou2019cross}, a cross attention module (CAN) was proposed to enhance the overall discrimination of few-shot classification \cite{sung2018learning}. Inspired by the human behavior of recognizing novel images, the similarity between seen and unseen parts is identified first. CAN learns to encode the correlation between the query and the target object. As Figure~\ref{fig:multimodals} (a) shows, the features of query and target are extracted independently, and then a correlation layer computes the interaction between them using cosine distance. Next, 1D convolution is applied to fuse correlation (GAP is performed first) and attentions, followed by softmax normalization. The output is reshaped to give a single-channel feature map to preserve spatial representations. Although experiments show that CAN produces state-of-the-art results, it depends on non-learnable functions such as the cosine correlation. Also, the design is suitable for few-shot classification but is not general because it depends on two streams (query and target).

\vspace{1.5mm} 
\noindent   
\textbf{Criss-Cross Attention}: The contextual information is still very important for scene understanding \cite{context1, context2}. Criss-cross attention proposed encoding the context of each pixel in the image in the criss-cross path. By building recurrent modules of criss-cross attention, the whole context is encoded for each pixel. This module is more efficient than non-local block \cite{m_nonlocal} in memory and time, where the memory is reduced by $11x$ and GFLOPS reduced by $85\%$. 
Since this survey focuses on the core attention
ideas, we show the criss-cross module 
in Figure~\ref{fig:multimodals} (b). Initially, three $1 \times 1$ convolutions are applied to the feature maps, whereas two of them are multiplied (the first map with each row of the second) to produce criss-cross attentions for each pixel. Then, softmax is applied to generate the attention scores, aggregated with the third convolution outcome. However, the encoded context captures only information in the criss-cross direction and not the whole image. For this reason,
the authors repeated the attention module by sharing the weights to form a recurrent criss-cross, which includes the whole context~\cite{huang2019ccnet}.

\vspace{1.5mm} 
\noindent   
\textbf{Perceiver Traditional}: CNNs have achieved high performance in handling several tasks \cite{resnet, chen2011multi, hassanin2021mitigating}, however, they are designed and trained for 
a single domain rather than multi-modal tasks \cite{modal1, modal2, modal3}. Inspired by biological systems that simultaneously understand the environment through various modalities, Jaegle~\etal proposed a perceiver that leverages the relations between these modalities iteratively. The main concept behind the perceiver is to form an attention bottleneck composed of a set of latent units. This solves the scale of quadratic processing, as in a traditional transformer, and encourages the model to focus on important features through iterative processing. To compensate for the spatial context, Fourier transform encodes the features \cite{mildenhall2020nerf, kandel2000principles, stanley2007compositional, parmar2018image}. As Figure~\ref{fig:multimodals} (c) shows, the perceiver is similar to RNN because of weight sharing. It composes two main components: cross attention to map the input image or input vector to a latent vector and transformer tower that maps the latent vector to a similar one with the same size. The architecture reveals that the perceiver is an attention bottleneck that learns a mapping function from high-dimensional data to a low-dimensional one and then passes it to the transformer \cite{m_transformers}. The cross-attention module has multi-byte attend layers to enrich the context, which might be limited from such mapping. This design reduces the quadratic processing $\mathcal{O}(M^2)$ to $\mathcal{O}(MN)$, where $M$ is the sequence length and $N$ is a hyperparameter that can be chosen smaller than $M$. Additionally, sharing the weights of the iterative attention reduces the parameters to one-tenth and enhances the model's generalization.

\vspace{1.5mm} 
\noindent   
\textbf{Stacked Cross Attention}: Lee~\etal~\cite{stacked_cross} proposed a method to attend between an image and a sentence context. Given an image and sentence, it learns the attention of words in a sentence for each region in the image and then scores the image regions by comparing each region to the sentence. This way of processing enables stacked cross attention to discover all possible alignments between text and image. Firstly, they compute image-text cross attention by a few steps as follows: a) compute cosine similarity for all image-text pairs \cite{karpathy2014deep} followed by $\ell_2$ normalization \cite{wang2017normface}, b) compute the weighted sum of these pairs attentions, where the image one is calculated by softmax \cite{chorowski2015attention}, c) the final similarity between these pairs is computed using LogSumExp pooling \cite{he2008discriminative, huang2018learning}. The same steps are repeated to get the text-image cross attention, but the attention in the second step uses text-based softmax. Although stacked attention enriches the semantics of multi-modal tasks by attending text over the image and vice versa, shared semantics might lead to misalignment in case of lack of similarity. With slight changes to the main concept, several works in various paradigms such as question answering and image captioning \cite{cross_attention1, cross_attention2, cross_attention4, cross_attention3, cross_attention5} used the stacked-cross attention.

\vspace{1.5mm} 
\noindent   
\textbf{Boosted Attention}: While top-down attention mechanisms~\cite{lu2017knowing} fail to focus on regions of interest without prior knowledge, visual stimuli methods~\cite{tavakoli2017paying,sugano2016seeing} alone are not sufficient to generate captions for images. For this reason, in~\ref{fig:multimodals} (d), the authors proposed a boosted attention model to combine both of them in one approach to focus on top-down signals from the language and attend to the salient regions from stimuli independently. Firstly, they integrate stimuli attention with visual feature $I^{'} = W\ I \circ \mathrm{log}(W_{sal}I+\epsilon))$, where $I$ is the extracted features from the backbone, $W_{sal}$ denotes the weight of the layer that produces stimuli attention, $W$ is the weight of the layer that output the visual feature layer. Boosted attention is achieved using the Hadamard product on $I^{'}$. Their experiments work showed that boosted attention improved performance significantly.

\subsubsection{Logical Attention}
\label{sec:logical}
Similar to how human beings pay more attention to crucial features, some methods have been proposed to use recurrences to encode better relationships. These methods rely on RNNs or any sequential network to calculate the attentions. We named it logical methods because they use architectures similar to logic gates.

\begin{figure}[tbp]
  \centering
  \begin{tabular}{cc}
    \includegraphics[width=.37\paperwidth]{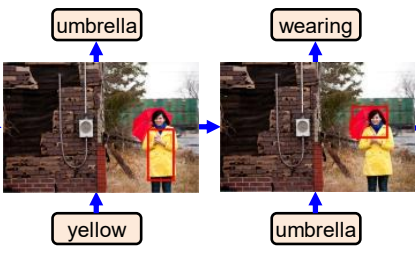} &     \includegraphics[width=.37\paperwidth]{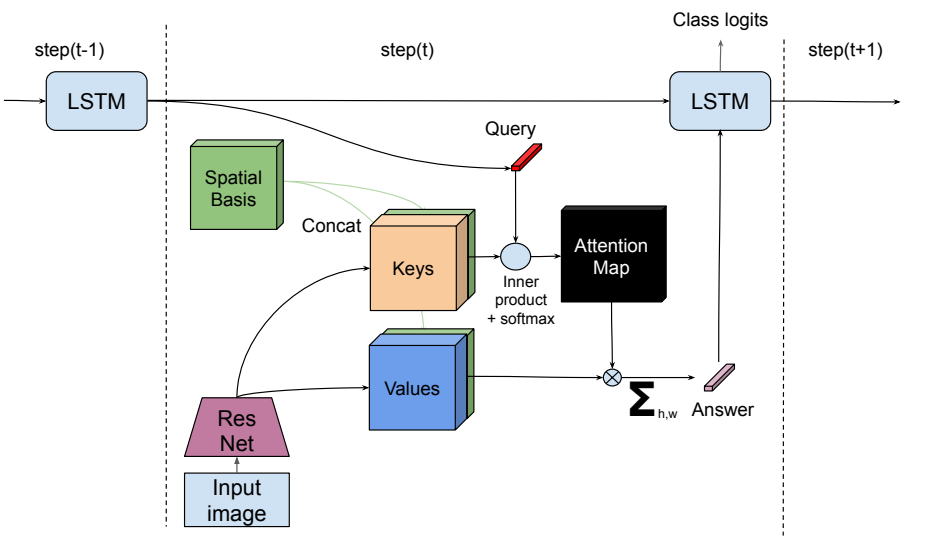}\\
    \small (a) Prophet ~ \cite{prophet}

&
    \small (b) S3TA~ \cite{zoran2020towards}
  \end{tabular}
   \caption{The core structure of logic-based attention methods such as Prophet attention~\cite{prophet} and S3TA  \cite{zoran2020towards} which are a type of attention that use logical networks such as RNN to infer the attention scores. Images are taken from the original papers and are best viewed in color.}
   \label{fig:logical}
\end{figure}

\vspace{1.5mm} 
\noindent   
\textbf{Sequential Attention Models}: Inspired by the primate visual system, Zoran~\etal~\cite{zoran2020towards} proposed soft, sequential, spatial top-down attention method (S3TA) to focus more on attended regions of an image \cite{mott2019towards} (as shown in Figure~\ref{fig:logical} (b)). At each step of the sequential process, the model queries the input and refines the total score based on spatial information in a top-down manner. Specifically, the backbone extracts features~\cite{resnet, huang2017densely, pham2018efficient} channels that are split into keys and values. A Fourier transform encodes the spatial information for these two sets to preserve the spatial information from disappearing for later use. The main module is a top-down controller, a version of the Long-Short Term Model (LSTM) \cite{lstm}, where its previous state is decoded as query vectors. The size of each query vector equals the sum of channels in keys and spatial basis. At each spatial location, the similarity between these vectors is calculated through the inner product, and then the softmax concludes the attention scores. These attention scores are multiplied by the values, and the summation is taken to produce the corresponding answer vector for each query. All these steps are in the current step of LSTM and are then passed to the next step. Note that the input of the attention module is an output of the LSTM state to focus more on the relevant information, as well as the attention map comprises only one channel to preserve the spatial information. Empirical evaluations show that attention is crucial for adversarial robustness because adversarial perturbations drag the object's attention away to degrade the model performance. Such an attention model proved its ability to resist strong attacks~\cite{madry2018towards} and natural noises \cite{hendrycks2019natural}. Although S3TA provides a novel method to empower the attention modules using recurrent networks, it is inefficient.

\vspace{1.5mm} 
\noindent   
\textbf{Permutation invariant Attention}: Initially, Zaheer~\etal~\cite{deep_sets} suggested handling deep networks in the form of sets rather than ordered lists of elements. For instance, performing pooling over sets of extracted features \eg $\rho(pool({\phi(x_1), \phi(x_2), \cdots, \phi(x_n)}))$,  where $\rho$ and $\phi$ are continuous functions and pool can be the $sum$ function. Formally, any set of deep learning features is an invariant permutation if $f(\pi x) = \pi f(x)$. Hence, Lee~\etal~\cite{permutation_invariant} proposed an attention-based method that processes sets of data. In \cite{deep_sets}, simple functions ($sum$ or $mean$) are proposed to combine the different branches of the network, but they lose important information due to squashing the data. To address these issues, set transformer~\cite{permutation_invariant} parameterizes the pooling functions and provides richer representations that can encode higher-order interaction. They introduced three main contributions: a) Set Attention Block (SAB), which is similar to Multi-head Attention Block (MAB) layer \cite{m_transformers}, but without positional encoding and dropout; b) induced Set Attention Blocks (ISAB), which reduced complexity from $\mathcal{O}(n^2)$ to $\mathcal{O}(mn)$, where $m$ is the size of induced point vectors and c) pooling by Multihead Attention (PMA) uses MAB over the learnable set of seed vectors.

\vspace{1.5mm}
\noindent
\textbf{Show, Attend and Tell}: Xu~\etal~\cite{xu2015show} introduced two types of attentions to attend to specific image regions for generating a sequence of captions aligned with the image using LSTM~\cite{zaremba2014recurrent}. They used two types of attention: hard attention and soft attention. Hard attention is applied to the latent variable after assigning multinoulli distribution to learn the likelihood of $log p(y|a)$, where $a$ is the latent variable. By using multinoulli distribution and reducing the variance of the estimator, they trained their model by maximizing a variational lower bound as pointed out in \cite{mnih2014recurrent, ba2015multiple} provided that attentions sum to $1$ at every point, \ie, $\sum_i \alpha_{ti} = 1$, where $\alpha$ refers to attentions scores. For soft attention, they used softmax to generate the attention scores, but for $p(s_t|a)$ as in \cite{baldi2014dropout}, where $s_t$ is the extracted feature at this step. The training of soft attention is easily done by normal back-propagation and to minimize the negative log-likelihood of $-\log(p(y|a))+\sum_i(1 - \sum_t\alpha_{ti})^2$. This model achieved the benchmark for visual captioning at the time as it paved the way for visual attention to progress.

\vspace{1.5mm} 
\noindent   
\textbf{Kalman Filtering Attention}: Liu~\etal identified two main limitations that hinder using attention in various fields where there is insufficient learning or history~\cite{kalman}. These issues are 1) the object's attention for input queries is covered by past training, and 2) conventional attentions does not encode hierarchical relationships between similar queries. To address these issues, they proposed the use of Kalman filter attention. Moreover, KFAtt-freq captures the homogeneity of the same queries, correcting the bias towards frequent queries.

\vspace{1.5mm} 
\noindent   
\textbf{Prophet Attention}: In prophet attention~\cite{prophet}, the authors noticed that the conventional attention models are biased and cause deviated focus to the tasks, especially in sequence models such as image captioning~\cite{captioning1, captioning2} and visual grounding~\cite{grounding1, grounding2}. Further, this deviation happens because attention models utilize previous input in a sequence to attend to the image rather than the outputs. As shown in Figure~\ref{fig:logical} (a), the model is attending on \enquote{yellow and umbrella} instead of \enquote{umbrella and wearing}. In a like-self-supervision way, they calculate the attention vectors based on the words generated in the future. Then, they guide the training process using these correct attentions, which can be considered a regularization of the whole model. Simply put, this method is based on summing the attentions of the post sequences in the same sentence to eliminate the impact of deviated focus toward the inputs. Overall, prophet attention addresses sequence-models biases towards history while disregarding the future.

\begin{figure}[t]
\centering
\includegraphics[width=.8\columnwidth]{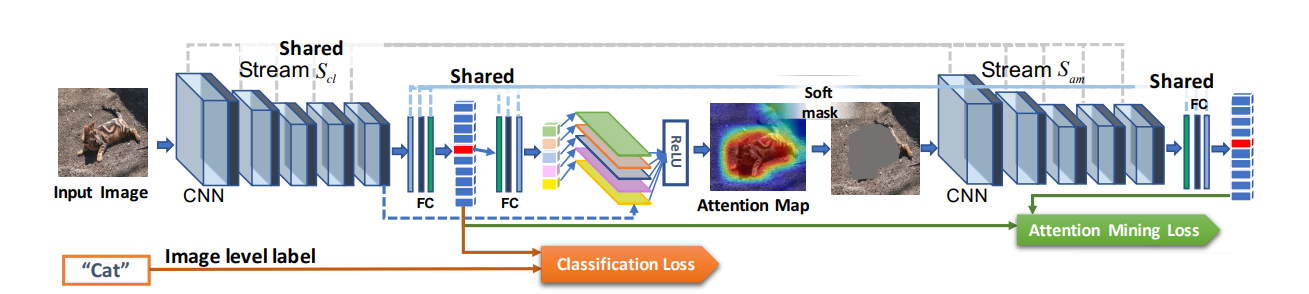}
\caption{An example of Guided Attention Inference Networks \cite{li2018tell}.}
\label{fig:class}
\end{figure}

\subsubsection{Category-Based Attentions}
\label{sec:Category}
The above methods generate the attention scores from the features regardless of the presence of the class. On the other hand, some techniques use class annotation to force the network to attend over specific regions. 

\vspace{1.5mm} 
\noindent   
\textbf{Guided Attention Inference Network}: In \cite{li2018tell}, the authors proposed class-aware attention, namely Guided Attention Inference Networks (GAIN), guided by the labels. Instead of focusing only on the most discriminative parts in the image~\cite{zhou2016learning}, GAIN includes the contextual information in the feature maps. Following~\cite{selvaraju2017grad}, GAIN obtains the attention maps from an inference branch, which are then used for training. As shown in Figure~\ref{fig:class}, through 2D-convolutions, global average pooling, and ReLU, the important features are extracted $A^c$ for each class. Following this, the features of each class are obtained as $I - (T(A^c)\bigodot I)$ where $\bigodot$ is matrix multiplication. $T(A^c) = \frac{1}{1+exp(-w(A^c - \sigma))}$ where $\sigma$ is a threshold parameter and $w$ is a scaling parameter. Their experiments showed that GAINS improved significantly over the state-of-the-art without recursive runs.

\vspace{1.5mm} 
\noindent  
\textbf{Curriculum Enhanced Supervised Attention Network}: The majority of attention methods are trained in a weakly supervised manner, and hence, the attention scores are still far from the best representations~\cite{m_transformers, m_nonlocal}. In~\cite{zhu2020curriculum}, the authors introduced a novel idea to generate a Supervised-Attention Network (SAN). Using the convolution layers, they defined the output of the last layer to be equal to the number of classes. Therefore, performing attention using global average pooling \cite{lin2013network} yields a weight for each category. In a similar study, Fukui~\etal proposed using a network composed of three branches to obtain class-specific attention scores: feature extractor to learn the discriminative features, attention branch to compute the attention scores based on a response model, perception to output the attention scores of each class by using the first two modules. The main objective was to increase the visual explanations \cite{zhou2016learning} of the CNN networks as it showed significant improvements in various fields such as fine-grained recognition and image classification.

\vspace{1.5mm} 
\noindent   
\textbf{Attentional Class Feature Network}:
Zhang~\etal~\cite{zhang2019acfnet} introduced ACFNet, a novel idea to exploit contextual information for improving semantic segmentation. Unlike conventional methods that learn spatial-based global information \cite{chen2017rethinking}, this contextual information is categorial-based, firstly presenting the class-center concept and then employing it to aggregate all the corresponding pixels to form a specific class representation. In the training phase, ground-truth labels are utilized to learn class centers, while coarse segmentation results are used in the test phase. Finally, class-attention maps are the results of class centers and coarse segmentation outcomes. The results show significant improvement in semantic segmentation using ACFNet.

\subsection{Hard (Stochastic) Attention}
Instead of using the weighted average of the hidden states, hard attention selects one of the states as the attention score. Proposing hard attention depends on answering two questions: (1) how to model the problem and (2) how to train it without vanishing the gradients. In this part, hard attention methods are discussed, as well as their training mechanisms. It includes a discussion of Bayesian attention, variational inference, reinforced, and Gaussian attentions. The main idea of Bayesian and variational attention is to use latent random variables as attention scores. Reinforced attention replaces softmax with a  Bernoulli-sigmoid unit \cite{williams1992simple}, whereas Gaussian attention uses a 2D Gaussian kernel instead. Similarly, self-critic attention~\cite{chen2019self} employs a re-enforcement technique to generate the attention scores, whereas Expectation-Maximization uses EM to generate the scores.

\subsubsection{Statistical-based attention}
\label{sec:statistical}
\begin{figure}
  \centering
  \begin{tabular}[b]{c}
   \includegraphics[width=.36\paperwidth]{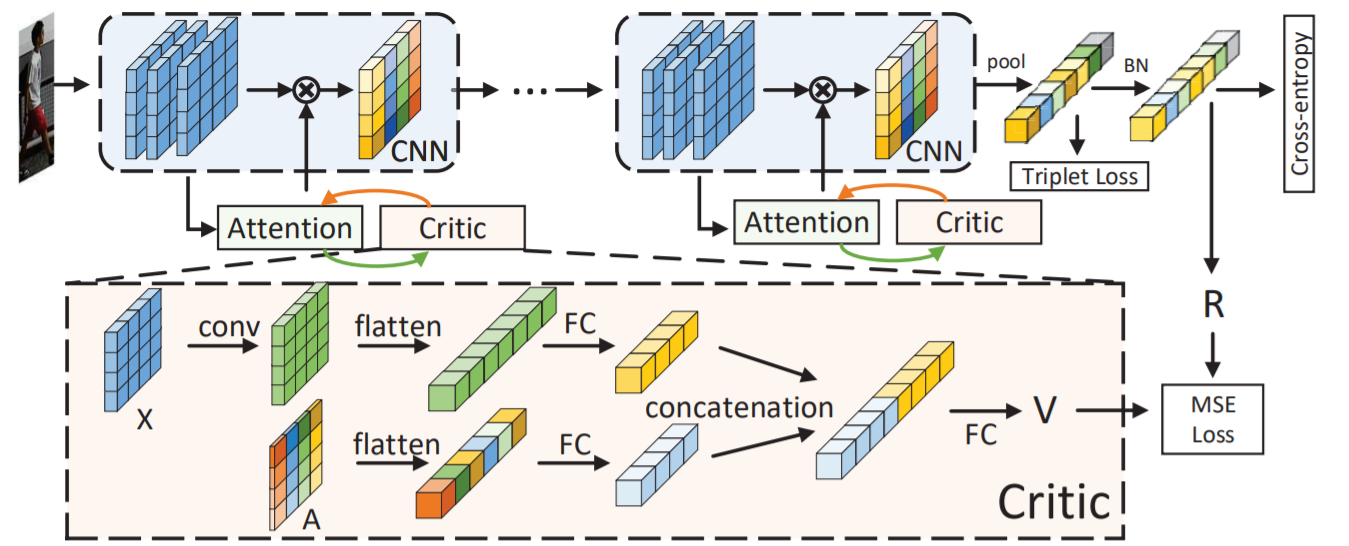} \\
    \small (a) Self-Critic Attention~\cite{chen2019self}\\
    \includegraphics[width=.36\paperwidth]{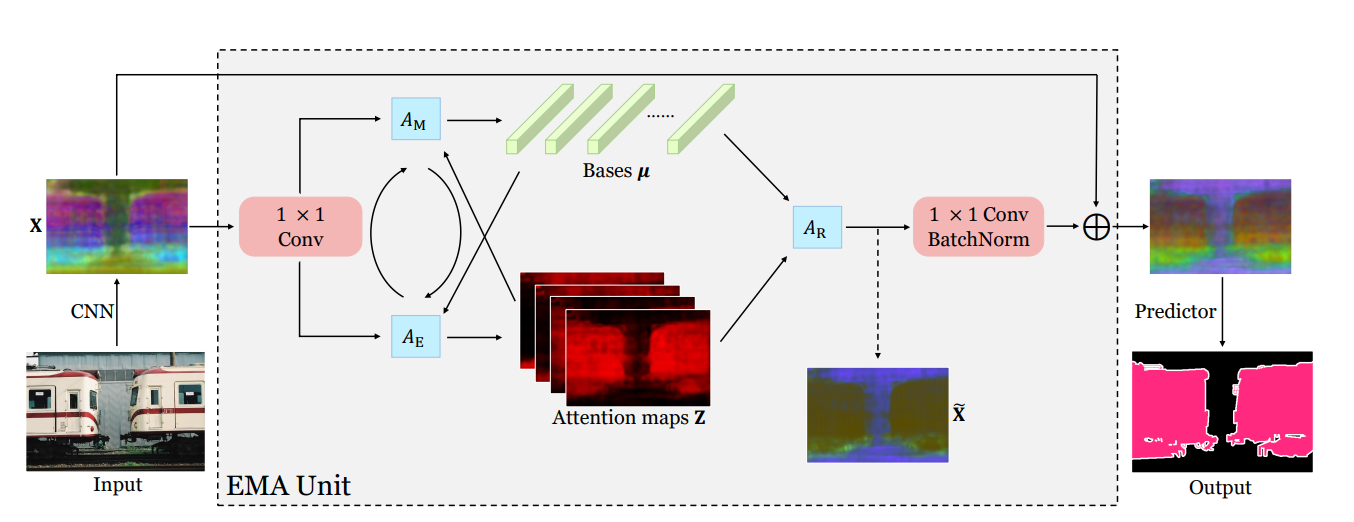} \\
    \small (c) Expectation-Maximization Attention ~ \cite{li2019expectation}
  \end{tabular}
  \begin{tabular}[b]{c}
  \includegraphics[width=.3\paperwidth]{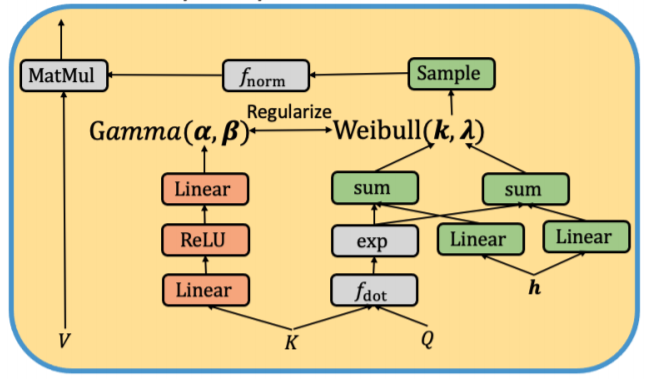}\\
   \small (b) Bayesian Attention Belief Networks \cite{zhang2021bayesian}
    \tabularnewline\\
   \includegraphics[width=.3\paperwidth]{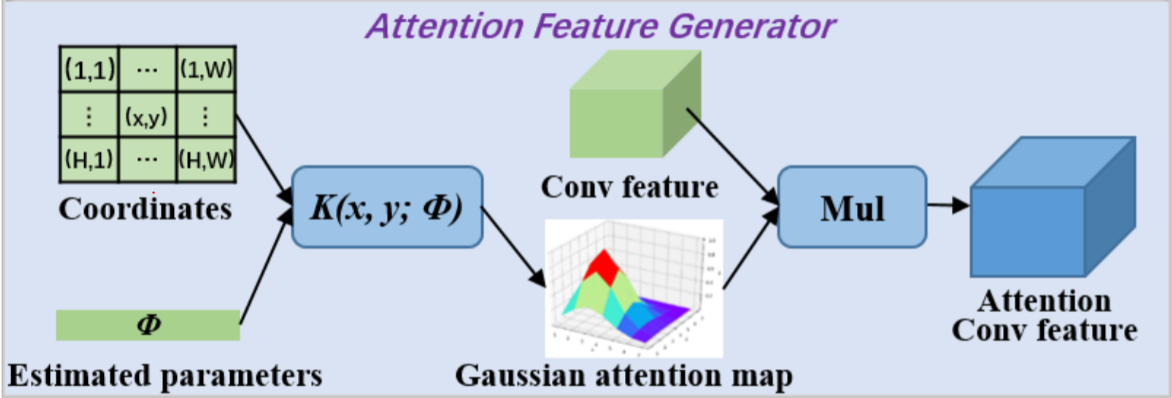} \\
    \small (d) Gaussian Attention~\cite{gaussian_attention}
  \tabularnewline
  \end{tabular}
   \caption{Illustration of hard attention architectures. Building blocks of EMA~\cite{li2019expectation}, Gaussian~\cite{gaussian_attention}, Self-critic~\cite{chen2019self} and Bayesian~\cite{zhang2021bayesian}. Images are taken from the original papers.}
     \label{fig:stochastics}
\end{figure}

\vspace{1.5mm} 
\textbf{Bayesian Attention Modules (BAM)}
In contrast to the deterministic attention modules, Fan~\etal~\cite{fan2020bayesian} proposed a stochastic attention method based on Bayesian-graph models. Firstly, keys and queries are aligned to form distribution parameters for attention weights, treated as latent random variables. They trained the whole model by reparameterization, which results from weight normalization by Lognormal or Weibull distributions. Kullback–Leibler (KL) divergence is used as a regularizer to introduce contextual prior distribution in the form of keys' functions. Their experiments illustrate that BAM significantly outperforms the state-of-the-art in various fields such as visual question answering, image captioning, and machine translation. However, this improvement happens on account of computational cost as well as memory usage. Compared to deterministic attention models, it is generally an efficient alternative, showing consistent effectiveness in language-vision tasks. 
\begin{itemize}

\vspace{1.5mm} 
\item \textbf{Bayesian Attention Belief Networks}: Zhang~\etal~\cite{zhang2021bayesian} proposed using Bayesian Belief modules to generate attention scores given their ability to model highly structured data along with uncertainty estimations. As shown in~\ref{fig:stochastics} (b), they introduced a simple structure to change any deterministic attention model into a stochastic one through four steps: 1) using Gamma distributions to build the decoder network 2) using Weibull distributions along with stochastic and deterministic paths for downward and upward, respectively 3) Parameterizing BABN distributions from the queries and keys of the current network 4) using evidence lower bound to optimize the encoder and decoder. The whole network is differentiable because of Weibull distributions in the encoder. Regarding accuracy and uncertainty, BABN proved improvement over the state-of-the-art in NLP tasks.

\vspace{1.5mm}  
\item \textbf{Repulsive Attention}: Multi-head attention \cite{m_transformers} is the core of attention used in transformers. However, MHA may cause attention collapse when extracting the same features \cite{an2020repulsive, prakash2019repr, han2016dsd} and consequently, the discrimination power for feature representation will not be diverse. To address this issue, An~\etal~\cite{an2020repulsive} adapted MHA to a Bayesian network with underlying stochastic attention. MHA is considered a special case without sharing parameters, and using a particle-optimization sample to perform Bayesian inference on the attention parameter imposes attention repulsiveness \cite{liu2016stein}. Through this sampling method, each MHA is considered a sample seeking posterior distribution approximation, far from other heads.
\end{itemize}

\vspace{1.5mm}
\noindent
\textbf{Variational Attention}
\label{sec:variational}
In a study to improve the latent variable alignments, Deng~\etal~\cite{deng2018latent} proposed using a variational attention mechanism. A latent variable is crucial because it encodes the dependencies between entities, whereas variational inference methods represent it in a stochastic manner \cite{salimbeni2019deep, drori2020deep}. On the other hand, soft attention can encode alignments, but it has poor representation because of the nature of softmax. Stochastic methods show better performance when optimized well~\cite{lin2003toward, wang2020survey}. The main idea is to propose variational attention while keeping the training tractable. They introduced two types of variational attention: categorical (hard) attention that uses amortized variational inference based on policy gradient and soft attention variance, and relaxed (probabilistic soft attention) using Dirichlet distribution that allows attending over multiple sources. Regarding reparameterization, Dirichlet distribution is not parameterizable; thus, the gradient has high variances~\cite{jankowiak2018pathwise}. Inspired by~\cite{deng2018latent}, Bahuleyan~\etal developed stochastic attention-based variational inference \cite{bahuleyan2018variational} but using a normal distribution instead of Dirichlet distribution. They observed that variational encoder-decoders should not have a direct connection; otherwise, traditional attentions serve as bypass connections.

\subsubsection{Reinforcement-based Attention}
\label{sec:reinforced} 

\vspace{1.5mm} 
\noindent   
~\\\textbf{Self-Critic Attention}:
\label{sec:self-critic}
Chen~\etal \cite{chen2019self} proposed a self-critic attention model that generates attention using an agent and re-evaluates the gain from this attention using the REINFORCE algorithm. They observed that most of the attention modules are trained in a weakly-supervised manner. Therefore, the attention maps are not always discriminative and lack supervisory signals during training \cite{lee2015deeply}. To supervise the generation of attention maps, they used a reinforcement algorithm to guide the whole process. As shown in Figure~\ref{fig:stochastics} (a), the feature maps are evaluated to predict whether they need self-correctness.

\vspace{1.5mm} 
\noindent 
\textbf{Reinforced Self-Attention Network}:
Shen~\etal~\cite{shen2018reinforced} used a reinforced technique to combine soft and hard attention in one method. Soft attention has shown effectiveness in modeling local and global dependencies, which output from the dot-product similarity \cite{bahdanau2014neural}. However, soft attention is based on the softmax function that assigns values to each item, even the non-attended ones, which weakens the whole attention. On the other hand, hard attention \cite{show_attend} attends to important regions or tokens only and disregards others. Despite its importance to textual tasks, it is inefficient in terms of time and differentiability \cite{williams1992simple}. Shen~\etal~\cite{shen2018reinforced} used hard attention to extract rich information and then feed it into soft attention for further processing. Simultaneously, soft attention is used to reward hard attention and, hence, stabilize the training process. Specifically, they used hard attention to encode tokens from the input in parallel while 
combining it with soft attention \cite{shen2018disan} without any CNN/RNN modules (see Figure~\ref{fig:stochastics} (e)). In \cite{karianakis2018reinforced}, reinforcement attention was proposed to extract better temporal context from video. Specifically, this attention module uses Bernoulli-sigmoid unit \cite{williams1992simple}, a stochastic module. Thus, to train the whole system, the REINFORCE algorithm is employed to stabilize the gradients \cite{jankowiak2018pathwise}.
 
\subsubsection{Gaussian-based Attention}
\label{sec:Gaussian}

\vspace{1.5mm} 
\noindent   
~\\\textbf{Self Supervised Gaussian-Attention}: Most soft-attention models use softmax to predict the attention of feature maps \cite{m_transformers, image_transfomer, zhang2020resnest}, which suffers from various drawbacks. In~\cite{niu2020gatcluster}, Niu~\etal proposed replacing the classical softmax with a Gaussian attention module. As shown in Figure~\ref{fig:stochastics} (d), they build a 2D Gaussian kernel to generate attention maps instead of softmax $K = e(-\frac{1}{\alpha}(u - \mu)^T \sum^{-1} (u - \mu))$ for each individual element, where $u = [x, y]^T$, $\mu = [\mu_x, \mu^y]^T$. A fully connected layer passes the extracted features, and the Gaussian kernel predicts the attention scores. Using Gaussian kernels proved its effectiveness in discriminating the important features. Since it does not require further learning steps, such as fully connected layers or convolutions, this significantly reduces the number of parameters. As stochastic training models need careful designs because of SGD mismatching~\cite{stochastic1, stochastic2, stochastic3}, the Gaussian attention model developed binary classification loss that takes normalized logits to suppress the low scores and discriminate the high ones. This normalization uses a modified version of softmax, where the input is squared and divided by temperature value (\eg batch size).

\vspace{1.5mm} 
\noindent   
\textbf{Uncertainty-Aware Attention}: Since attention is generated without full supervision (\ie, in a weakly-supervised manner), it lacks full reliability \cite{li2018tell}. To fix this issue, \cite{heo2018uncertainty} proposed using uncertainty based on input. It generates varied attention maps according to the input and, therefore, learns higher variance for uncertain inputs. Gaussian distribution handles attention weights, giving small values in case of high confidence and vice versa \cite{kendall2017uncertainties}. Bayesian network is employed to build the model with variational inference as a solution~\cite{zhang1994simple, blei2017variational}. Note that this model is stochastic, and SGD back-propagation flow can not work properly due to randomness \cite{kingma2013auto}. For this reason, they used the reparameterization trick \cite{gal2017concrete, kingma2015variational} to train their model.

\subsubsection{Clustering}
\label{sec:EM}
~\\\textbf{Expectation Maximization attention}:
Traditional soft attention mechanisms can encode long-range dependencies by comparing each position to all positions, which is computationally very expensive~\cite{m_nonlocal}. In this regard, Li~\etal~\cite{li2019expectation} proposed using expectation maximization to build an attention method that iteratively forms a set of bases that compute the attention maps \cite{dempster1977maximum}. The main intuition is to use expectation maximization to select a compact basis set instead of using all the pixels as in \cite{m_nonlocal, li2020spatial} (see Figure~\ref{fig:stochastics} (c)). These bases are regarded as the learning parameters, whereas the latent variables serve as the attention maps. The output is the weighted sum of bases, and the attention maps are the weights. The estimation step is defined by $ z=\frac{\mathbb{K}(x_n, \mu_n)}{\sum_j\mathbb{K}(x_n, \mu_j)}$, where $\mathbb{K}$ denotes a kernel function. The maximization step updates $\mu$ through data likelihood maximization such that $\mu = \frac{z_{nk}(x_n, \mu_n)}{\sum_j z_{jk}}$. Finally, the features are multiplied by attention scores $\mathbb{X} = \mathbb{Z} \mu$. Since EMA is a stochastic model, training the whole model needs special care. Firstly, the authors average the $\mu$ over the mini-batch and update the maximization step to train it stably. Secondly, they normalize the Value of $\mu$ to be within (1, $T$) by $\ell_2$-Norm. EMA has shown the ability to remove noisy representation and to give promising results after the third step. Also, it is worth noting that the complexity is reduced to a linear form $\mathcal{O}(NK)$ from a quadratic one $\mathcal{O}(N^2)$.

%% file: table.tex
\renewcommand{\arraystretch}{1.2}

\begin{table*}
\caption{Summary of attention types along with their categorization, applications, strengths, and limitations. References to the original papers are also provided, and links to sections where they are discussed.}
\centering
\resizebox{\textwidth}{!}{
\small
\begin{tabular}
{>{\raggedright}m{0.01\textwidth}>{\raggedright}m{0.01\textwidth}>{\raggedright}m{0.01\textwidth}>{\raggedright}m{0.22\textwidth}>{\raggedright}m{0.22\textwidth}>{\raggedright}m{.24\textwidth}>{\raggedright}m{0.24\textwidth}}
\hline 
\rotatebox[origin=c]{90}{Type} & \rotatebox[origin=c]{90}{Category} & \rotatebox[origin=c]{90}{Section}& References & Applications & Strengths & Limitations\tabularnewline
\hline 
& \rotatebox[origin=c]{90}{Channel} & \rotatebox[origin=c]{90}{\ref{sec:channel}}&SE-Net~\cite{se}, ECA-Net~\cite{wang2020eca}, CBAM~\cite{woo2018cbam},~\cite{harmonious}, A2-Net~\cite{ding2020high}, Dual~\cite{fu2019dual}  & VR, Re-ID, medical segmentation
 & \textbullet\ easy to model \newline
     \textbullet\  differentiable gradients \newline
     \textbullet\  non-local operations \newline
      \textbullet\  encoding context    \newline
      \textbullet\  improving the performance
     
     & 
         \textbullet\   memory usage \newline 
         \textbullet\  computation cost \newline
         \textbullet\  radial softmax  \newline
          \textbullet\  lack of generalization
         
       \tabularnewline
\cline{4-5} 
 & \rotatebox[origin=c]{90}{\ \ Spatial\ \ } &\rotatebox[origin=c]{90}{\ref{sec:spatial}} &CBAM \cite{woo2018cbam}, PFA \cite{zhao2019pyramid}, \cite{li2020spatial}, \cite{meng2020end}  & VR, domain adaptation, saliency detection  & & \tabularnewline
\cline{4-5}
 & \rotatebox[origin=c]{90}{Self-}&\rotatebox[origin=c]{90}{\ref{sec:self}} & ViT \cite{m_transformers}, Image ViT\cite{image_transfomer}, \cite{m_self_attention}, \cite{m_standalone}  & VR, multi-modals, low-levels, 3D analysis & &\tabularnewline
\cline{4-5}

 \multirow{9}{*}{{\rotatebox[origin=c]{90}{\textbf{Soft (Deterministic)}}}} &\rotatebox[origin=c]{90}{Category} & \rotatebox[origin=c]{90}{\ref{sec:Category}} & GAIN~\cite{li2018tell}~\cite{zhu2020curriculum}  & explainable machine learning, Re-ID, semantic segmentation&   
     \textbullet\  gradient understanding \newline
     \textbullet\   no extra supervision 
      &
     \textbullet\  extra computation \newline
     \textbullet\  only supervised classification
      \tabularnewline
\cline{4-5}

&\rotatebox[origin=c]{90}{Multi-modal} &\rotatebox[origin=c]{90}{\ref{sec:Multi-modal}}& CAN~\cite{hou2019cross}, SCAN~\cite{stacked_cross}, Perceiver~\cite{jaegle2021perceiver}, Boosted~\cite{chen2018boosted} & few-shot, image-text matching, captioning &  
     \textbullet\  multi-modal applications\newline  
     \textbullet\  supervision signals \newline
     \textbullet\   accuracy rates 
      &
     \textbullet\  memory usage \newline
     \textbullet\  computation cost \newline
     \textbullet\  soft \& hard issues 
        \tabularnewline

\cline{4-5}
 
&\rotatebox[origin=c]{90}{Arithmetic} & \rotatebox[origin=c]{90}{\ref{sec:arithmetic}}& Drop-out \cite{baldi2014dropout}, Mirror~\cite{jin2020semantic}, Reverse~\cite{chen2018reverse}, Inverse~\cite{zhang2020robust},  Reciprocal~\cite{xia2019exploring}  & weakly detection, detection and  segmentation,   &  
     \textbullet\  efficient methods \newline
     \textbullet\  simple ideas \newline
     \textbullet\  easy to implement\newline
     \textbullet\  semantics of the models
      &
     \textbullet\  certain applications \newline 
     \textbullet\  inability to scale up \newline
     \textbullet\  limitations of soft and hard  
      \tabularnewline
\cline{4-5}
 
&\rotatebox[origin=c]{90}{Logical} &\rotatebox[origin=c]{90}{\ref{sec:logical}}&Recurrent~\cite{liu2018picanet}, Sequential~\cite{zoran2020towards, NEURIPS2020_103303dd}, Permutation invariant~\cite{permutation_invariant} & VR, dense prediction, adversarial classification,  tagging, anomaly detection    &  
     \textbullet\  overcoming soft-attention issues\newline
     \textbullet\  addressing hard drawbacks
      &
     \textbullet\  complex architectures\newline
     \textbullet\  high computation cost \newline
     \textbullet\  iterative processing
      \tabularnewline
\hline

 & \rotatebox[origin=c]{90}{Statistical}&  \rotatebox[origin=c]{90}{\ref{sec:statistical}}& Bayesian~\cite{fan2020bayesian}, Repulsive~\cite{an2020repulsive}, Variational \cite{deng2018latent},~\cite{xu2015show}  & VQA, captioning, image translation, NLP &  & \tabularnewline
\cline{4-5}
\multirow{6}{*}{{\rotatebox[origin=c]{90}{\textbf{Hard (Stochastic)}}}}& \rotatebox[origin=c]{90}{Reinforced} & \rotatebox[origin=c]{90}{\ref{sec:reinforced}} &RESA~\cite{shen2018reinforced},~\cite{karianakis2018reinforced}, self-critic \cite{chen2019self} & Re-ID, NLP  & 
     \textbullet\   context \newline
     \textbullet\   higher-order interactions \newline
     \textbullet\  diverse attention scores \newline
     \textbullet\  higher improvements
      & 
     \textbullet\  memory usage \newline
     \textbullet\  computation cost \newline
     \textbullet\  non-differentiable \newline
     \textbullet\  gradient vanishing\newline
     \textbullet\  tricks for training
     \tabularnewline
\cline{4-5}
 & \rotatebox[origin=c]{90}{\ Gaussian\ }& \rotatebox[origin=c]{90}{\ref{sec:Gaussian}}& GatCluster~\cite{gaussian_attention}, Uncertainty~\cite{heo2018uncertainty}  & clustering  medical NLP   && \tabularnewline
\cline{4-5}

 & \rotatebox[origin=c]{90}{\ Clustering\ } &\rotatebox[origin=c]{90}{\ref{sec:EM}}&EM~\cite{li2019expectation}, GatCluster~\cite{gaussian_attention} & semantic segmentation &  & \tabularnewline
\hline 

\end{tabular}
}
\label{tab:overall}
\end{table*}

%% file: AttAppType.tex
\section{Attention for Vision Tasks}
\label{sec:AttAppType}
As mentioned above, attention mechanisms have become critical in various computer vision tasks, allowing models to focus on specific parts of an image. Based on our survey, we deduce a general trend where one type of attention mechanism is better suited to a particular computer vision application than the others. Following are some of the examples,

\begin{enumerate}
\item \textit{Spatial Attention} is commonly used in tasks like image captioning, where the model learns to attend to relevant regions while generating descriptions.

\item \textit{Channel Attention} helps enhance the importance of informative channels while suppressing less relevant ones, leading to improved feature representation. It is used in tasks like image classification and object detection.

\item \textit{Spatio-Temporal Attention} extends the concept to video data, allowing models to attend to specific spatial regions and temporal frames. It is valuable in video analysis tasks, such as action recognition and video captioning

\item \textit{Self-Attention Mechanism} is used in transformers and has revolutionized various computer vision tasks, including image captioning, object detection, and image generation.

\item \textit{Cross-Modal Attention} allows the model to align and attend to relevant information between different modalities, enabling tasks like image-text matching and visual question answering.

\item \textit{Multi-Head Attention} enhances the representation power and has been applied in complex computer vision tasks like image segmentation and generative modeling.

\item \textit{Non-Local Attention} has been used in video processing tasks for capturing global contextual information across frames.

\item \textit{Graph Attention} is employed in graph-based computer vision tasks, such as graph-based image segmentation. Graph attention helps models focus on relevant nodes or edges in the graph structure, facilitating efficient and accurate processing.
\end{enumerate}

%% file: challenges.tex
\section{Open Problems and Challenges}
Despite the performance improvement and interesting salient features of attention models, various challenges are associated with their practical settings in computer vision applications. The essential impediments include a requirement for high computational costs, significant amounts of training data, the efficiency of the model, and a cost-benefit analysis of performance improvement. In addition, there have also been some challenges in visualizing and interpreting attention blocks. This section provides an overview of these challenges and limitations, mentions recent efforts to address those limitations, and highlights the open research questions.

\vspace{1mm} 
\noindent   
\textbf{Generalization:}  Attention models' generalization is a challenging task. Many of the proposed models are specific to the application underhand and only work well in the proposed settings. Whereas some models (\eg channel and spatial attention) have performed better in classification since attention models are primarily designed for high-level tasks, they fail when applied directly to low-level vision tasks. Moreover, the data quality notably influences the generalization and robustness of attention models. Thus, there is still a significant step to generalize pre-trained attention models on more generalized low-level vision tasks. Probably, the incorporation of various attention mechanisms within a multi-head type of framework offers a promising door to tackle this issue.   

\vspace{1mm} 
\noindent   
\textbf{Efficiency:} Efficiency of vision models is vital for many real-time computer vision applications. Unfortunately, current models focus more on performance than efficiency. Recently, self-attention has been successfully applied in transformers and shown to achieve better performance; however, at the cost of huge computational complexity \eg the base ViT~\cite{dosovitskiy2020image} has 18 billion FLOPs compared to the CNN models~\cite{han2020ghostnet,anwar2019real} with  600 million FLOPs, achieving similar performance to process an image. Although fewer attempts, such as Efficient Channel attention, are made to make the attention models more efficient, they remain complex to train; hence, efficient models are required for deployment on real-time devices.

\vspace{1mm} 
\noindent   
\textbf{Multi-Model Data:} Attention has been applied mainly on single domain data and in a single task setting. An important question is whether the attention model can fuse input data meaningfully and exploit multiple label types (or tasks) in the data. Also, it is yet to be seen that attention models can leverage the various labels available, such as combining the point clouds and the RGB images of the KITTI dataset, to provide a meaningful performance. Similarly, attention models can also be used to know whether they can predict relationships between the labels, actions, and attributes in a unified manner. In this context, a potential resolution might involve integrating these attention mechanisms into graph attention networks, thus constructing adept models for heterogeneous input data.

\vspace{1mm} 
\noindent   
\textbf{Amount of Training Data:} Attention models usually rely on more training data to learn the important aspects than simple non-attentional models. For example, self-attention employed in transformers needs to learn the invariance, translation, \etc by themselves instead of non-attentional CNNs, where these properties are inbuilt due to operations such as pooling. The increase in data also means more training time and computational resources. Hence, an open question is how to address this problem with more efficient attention models. 

\vspace{1mm} 
\noindent   
\textbf{Performance Comparisons:} Models that employ attentional blocks mostly compare their performance against the baseline without having the attention while ignoring other attentional blocks. The lack of comparison between different attentional models provides little information about performance improvement against other attentions. Therefore,  there is a need to present a more in-depth analysis of the increased parameters versus the performance gain of different attentional models proposed in the literature. 

\vspace{1mm} 
\noindent 
\textbf{Visualization Complexity:} Due to the intrinsic complexity of attention mechanisms involving several layers and operations, it might be challenging to comprehend how various features are weighted and what factors significantly influence the module's output. Attention modules are often treated as black boxes because of the number of parameters involved. Although attention visualization can provide some insights into attention module behavior, it may not fully illustrate the underlying weighing process, leading to an incomplete interpretation. In some attention methods, the attention weights might be very sparse; hence, it is difficult to determine meaningful patterns from the attention distributions. Moreover, the interpretation of attention is highly context-dependent and different input features can vary depending on the specific task.

\vspace{1mm} 
\noindent 
\textbf{Lack of Attentional Visualization $\&$ Interpretation:} Input to the attention modules are high dimensional features, especially in large models with many layers, making it difficult to visualize and interpret the attention patterns effectively, as inspecting every single attention weight becomes impractical. While limited visualization techniques exist for attention mechanisms; however, a single interpretation strategy may not be applicable to all attention modules. Moreover, it is also possible that different visualization methods may highlight various aspects of attention but may lead to potentially conflicting interpretations. Without oversimplification or distortion, standard visualization techniques that effectively capture the relevant information are ongoing challenges. Furthermore, many computer vision tasks process multiple modalities and developing attention visualization mechanisms that can effectively demonstrate the contribution of each modality while preserving meaningful relationships presents a significant challenge.

%% file: conclusions.tex
\section{Conclusions}
This paper reviewed more than 70 articles related to various attention mechanisms used in vision applications. We comprehensively discussed the attention techniques and their strengths and limitations. We offered to restructure the existing attention mechanisms proposed in the literature into a hierarchical framework based on how they compute their attention scores. Choosing the attention score calculation to group the reviewed techniques has effectively determined how the attention-based models are built and which training strategies are employed therein.

Although the capability of the developed attention-based techniques in modeling the salient features and boosting the performance is commendable, various challenges and open questions still need to be answered, especially when employing these techniques for computer vision tasks. We have listed these challenges and highlighted research questions that remain open. Despite some recent efforts introduced to cope with these limitations, we still need to solve attention-related problems in vision. This survey will help researchers focus on addressing these challenges efficiently and develop attention mechanisms better suited for vision-based applications.

\section{Acknowledgments}
Professor Ajmal Mian is the recipient of an Australian Research Council Future Fellowship Award (project number FT210100268) funded by the Australian Government. We thank Professor Mubarak Shah for his valuable comments that significantly improved the survey presentation.